%% file: root.tex
\documentclass[conference]{IEEEtran} 

\input{preamble.tex}

\input{glossary.tex}

\input{notation.tex}

\begin{document}
\title{Fast Traversability Estimation for \\ Wild Visual Navigation}

\author{Jonas Frey$^{\star,1,3}$ \hspace{4pt} Matias Mattamala$^{\star,2}$ \hspace{4pt} Nived Chebrolu$^2$ \hspace{4pt} Cesar Cadena$^1$ \hspace{4pt} Maurice Fallon$^2$ \hspace{4pt} Marco Hutter$^1$\\[4pt]
$^1$\,ETH Zurich \hspace{12pt} $^2$\,University of Oxford \hspace{12pt} $^3$\,Max Planck Institute for Intelligent Systems\\[2pt]
$^\star$\,Equal contribution. \texttt{jonfrey@ethz.ch}, \texttt{matias@robots.ox.ac.uk}
}

\twocolumn[{%
\renewcommand\twocolumn[1][]{#1}%
\maketitle
\begin{center}
    \vspace{-5pt}
    \centering
	\includegraphics[width=1.0\textwidth]{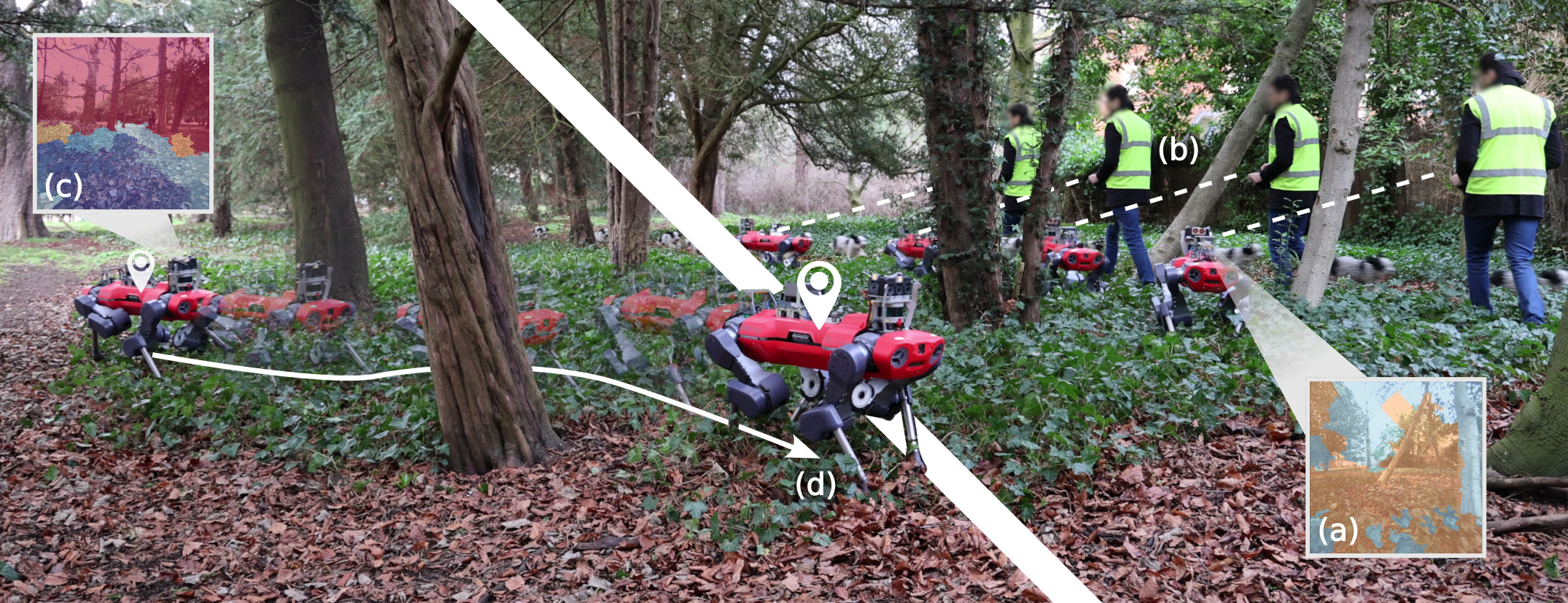}
	\captionof{figure}{\gls{wvn} learns to predict traversability from images via online self-supervised learning. Starting from a randomly initialized traversability estimation network without prior assumptions about the environment (a), a human operator drives the robot around areas that are traversable for the given platform (b). After a few minutes of operation, \gls{wvn} learns to distinguish between traversable and untraversable areas (c), enabling the robot to navigate autonomously and safely within the environment (d).}
    \label{fig:header}
\end{center}%
}]


%
\begin{abstract}
\input{chapters/0_abstract}
\end{abstract}
%
%
%

\section{Introduction}
\label{sec:Introduction}
\input{chapters/1_introduction}

\section{Related Work}
\label{sec:Related work}
\input{chapters/2_related_work}

\section{Method}
\label{sec:Method}
\input{chapters/3_method}

\section{Closed-loop Integration}
\label{sec:closed-loop_integration}
\input{chapters/4_closed_loop_integration.tex}

\section{Experiments}
\label{sec:Experiments}
\input{chapters/5_experiments}

\section{Conclusion}
\label{sec:Conclusion}
\input{chapters/6_conclusion}

\section*{Acknowledgments}
\label{sec:acknowledgments}
\input{chapters/acknowledgments}

\balance
\bibliographystyle{plainnat}
\bibliography{bibliography}

\vfill


\end{document}

%% file: preamble.tex
\usepackage{times}
\usepackage{amsmath,amsfonts}
\usepackage{algorithmic}
\usepackage{algorithm}
\usepackage{array}
\usepackage{balance}
\usepackage[dvipsnames,table]{xcolor}
\usepackage{textcomp}
\usepackage{stfloats}
\usepackage{url}
\usepackage{verbatim}
\usepackage{graphicx}
\usepackage{todonotes}
\usepackage{booktabs} 
\usepackage{bbm} 
\usepackage{mathtools}
\usepackage{pgf} 
\usepackage{etoolbox} 

\usepackage[numbers]{natbib} 

\usepackage{multicol}
\usepackage[bookmarks=true, backref]{hyperref} 
\usepackage{lipsum} 
\usepackage{siunitx} 
\usepackage{mathtools}
\usepackage{xspace}
\usepackage{annotate-equations} 

\usepackage{multirow}
\usepackage{pbox}

\usepackage{threeparttable}
\usepackage{tablefootnote}

\usepackage{tikz}

\usepackage[normalem]{ulem}
\definecolor{MK_Two_One}{RGB}{178,24,43} 
\definecolor{MK_Two_Two}{RGB}{239,138,98}
\definecolor{MK_Two_Three}{RGB}{253,219,199}
\definecolor{MK_Two_Four}{RGB}{209,229,240}
\definecolor{MK_Two_Five}{RGB}{103,169,207}
\definecolor{MK_Two_Six}{RGB}{33,102,172} 

\hypersetup{
colorlinks=true
,linkcolor=MK_Two_Six
,citecolor=MK_Two_Six
,filecolor=MK_Two_Six
,urlcolor= MK_Two_Six
,menucolor=MK_Two_Five
,runcolor=MK_Two_Four
,linkbordercolor=MK_Two_One
,citebordercolor=MK_Two_Two
,filebordercolor=MK_Two_Three
,urlbordercolor=MK_Two_Six
,menubordercolor=MK_Two_Five
,runbordercolor=MK_Two_Four
}

\usepackage{subcaption}
\definecolor{high}{RGB}{116, 173, 209}  
\definecolor{low}{RGB}{244, 109, 67}  

\newcommand{\ra}[1]{\renewcommand{\arraystretch}{#1}}

\def\secref#1{Sec.~\ref{#1}}

\def\figref#1{Fig.~\ref{#1}}
\def\tabref#1{Tab.~\ref{#1}}
\def\eqref#1{Eq.~(\ref{#1})}

\usepackage{tikz}


%% file: glossary.tex
\usepackage[abbreviations]{glossaries-extra}

\glssetcategoryattribute{abbreviation}{indexonlyfirst}{true}

\glssetcategoryattribute{abbreviation}{nohyperfirst}{true}

\newabbreviation{auroc}{AUROC}{Area Under the Receiver Operating Characteristic Curve}
\newabbreviation{accuracy}{Acc}{Accuracy}

\newabbreviation{cnn}{CNN}{Convolutional Neural Network}

\newabbreviation{fov}{FoV}{Field of View}
\newabbreviation{fpr}{FPR}{False Positive Ratio}

\newabbreviation{gnn}{GNN}{Graph Neural Network}
\newabbreviation{gcn}{GCN}{Graph Convolutional Network}
\newabbreviation{imu}{IMU}{Inertial Measurement Unit}
\newabbreviation{irl}{IRL}{Inverse Reinforcement Learning}

\newabbreviation{knn}{KNN}{K-Nearest Neighbors}

\newabbreviation{lagr}{LAGR}{Learning Applied to Ground Vehicles}
\newabbreviation{lidar}{LiDAR}{Light Detection and Ranging}

\newabbreviation{mlp}{MLP}{Multi-Layer Perceptron}
\newabbreviation{mpc}{MPC}{Model Predictive Controller}
\newabbreviation{mse}{MSE}{Mean Squared Error}

\newabbreviation{ood}{OOD}{out-of-distribution}

\newabbreviation{rbf}{RBF}{Radial Basis Function}
\newabbreviation{rmp}{RMP}{Riemannian Motion Policies}
\newabbreviation{ros}{ROS}{Robot Operating System}
\newabbreviation{ros1}{ROS~1}{Robot Operating System}
\newabbreviation{roc}{ROC}{Receiver Operating Characteristic}
\newabbreviation{rf}{RF}{Random Forest}

\newabbreviation{sdf}{SDF}{Signed Distance Field}
\newabbreviation{slam}{SLAM}{Simultaneous Localization and Mapping}
\newabbreviation{svm}{SVM}{Support Vector Machine}
\newabbreviation{svc}{SVC}{Support Vector Classifier}
\newabbreviation{wvn}{WVN}{Wild Visual Navigation}

\newabbreviation{vit}{ViT}{Vision Transformer}

%% file: notation.tex
\newcommand{\pose}[3]{\mathbf{T}_{\mathtt{#1 #2}}_{#3}}
\newcommand{\rot}[3]{\mathbf{R}_{\mathtt{#1 #2}}_{#3}}
\newcommand{\pos}[2]{{\mathtt{_#1}} \mathbf{p}_{#2}}

\newcommand{\vel}[1]{\mathbf{v}_{#1}}
\newcommand{\velcmd}[1]{\mathbf{\bar{v}}_{#1}}

\newcommand{\speed}[1]{v_{#1}}
\newcommand{\speedcmd}[1]{\bar{v}_{#1}}

\newcommand{\veldiff}{v_{\text{error}}}

\newcommand{\K}{\mathbf{K}_{3\times3}}

\newcommand{\img}[1]{\mathbf{I}^{#1}}

\newcommand{\feat}[1]{\mathbf{F}^{#1}}

\newcommand{\embed}[1]{\mathbf{f}_{#1}}

\newcommand{\segmask}[1]{\mathbf{M}^{#1}}

\newcommand{\auxsupimage}[1]{\mathbf{S}^{#1}}

\newcommand{\nsupg}{N_{\mathrm{sup}}}
\newcommand{\dsupg}{d_{\mathrm{sup}}}

\newcommand{\dmisg}{d_{\mathrm{mis}}}

\newcommand{\loss}[1]{\mathcal{L}_{\mathrm{#1}}}

\newcommand{\sigmafactor}{k_{\sigma}}
\newcommand{\sigmaanomi}{\sigma_{\mathrm{pos}}}
\newcommand{\muanomi}{\mu_{\mathrm{pos}}}

\newcommand{\SEtwo}{\mathrm{SE(2)}}
\newcommand{\SEthree}{\mathrm{SE(3)}}

\newcommand{\fun}[2]{f_{\mathrm{#1}}\left( #2 \right) }
\newcommand{\sigmoid}[2]{\mathrm{sigmoid}_{\mathrm{#1}}\left( #2 \right) }
\newcommand{\dimss}[2]{#1 \times #2}
\newcommand{\dimsss}[3]{#1 \times #2 \times #3}

\newcommand{\Rn}[1]{\mathbb{R}^{#1}}

\robustify{\pose}
\robustify{\rot}
\robustify{\pos}
\robustify{\K}
\robustify{\loss}
\robustify{\feat}
\robustify{\img}
\robustify{\fun}

\newcommand{\trav}[1]{{\tau}_{#1}}

\newcommand{\travthr}[0]{{\tau}_{\text{thr}}}

\newcommand{\dino}[1]{\text{DINO-ViT}{#1}}

\definecolor{TraversableBlue}{RGB}{49, 54, 149}
\definecolor{UntraversableRed}{RGB}{192, 26, 38}
\definecolor{PaperOrange}{RGB}{251, 151, 39}
\definecolor{PaperMagenta}{RGB}{150, 36, 145}
\definecolor{PaperBlue}{RGB}{67, 110, 176}
\definecolor{PaperCyan}{RGB}{66, 173, 187}

\usepackage{amssymb}
\newcommand{\travsquare}{{\textcolor{TraversableBlue}{$\blacksquare$}}}
\newcommand{\untravsquare}{{\textcolor{UntraversableRed}{$\blacksquare$}}}

\newcommand{\orangesquare}{{\textcolor{PaperOrange}{$\blacksquare$}}}
\newcommand{\magentasquare}{{\textcolor{PaperMagenta}{$\blacksquare$}}}

\newcommand{\cyansquare}{{\textcolor{PaperCyan}{$\blacksquare$}}}

%% file: chapters/0_abstract.tex
Natural environments such as forests and grasslands are challenging for robotic navigation because of the false perception of rigid obstacles from high grass, twigs, or bushes.
In this work, we propose \glsfirst{wvn}, an online self-supervised learning system for traversability estimation which uses only vision. The system is able to continuously adapt from a short human demonstration in the field. It leverages high-dimensional features from self-supervised visual transformer models, with an online scheme for supervision generation that runs in real-time on the robot. We demonstrate the advantages of our approach with experiments and ablation studies in challenging environments in forests, parks, and grasslands. Our system is able to bootstrap the traversable terrain segmentation in less than \SI{5}{\minute} of in-field training time, enabling the robot to navigate in complex outdoor terrains --- negotiating obstacles in high grass as well as a \SI{1.4}{\kilo\meter} footpath following. While our experiments were executed with a quadruped robot, ANYmal, the approach presented can generalize to any ground robot. Project~page:~\url{bit.ly/3M6nMHH}

%% file: chapters/1_introduction.tex
Traversability estimation is a core capability needed to allow robots to autonomous navigate in field environments. It is understood as the \emph{affordance}~\citep{Gibson1979} necessary for a robot to navigate within its environment, i.e to understand which areas can be accessed and navigated through and at what cost. While the topic has been widely studied for wheeled or flying robots supported by 3D sensors using the traditional approach of occupancy mapping~\citep{Moravec1985}, the development of new platforms with advanced mobility skills, such as legged robots, prompts a reconsideration of current definitions of traversability, as new and more complex types of natural terrain can be traversed~\citep{Miki2022a}.

It is difficult to infer traversability within natural terrains, such as high grass or forest undergrowth. Occupancy-based navigation systems based on 3D sensing are often confused by high grass and incorrectly classify such terrain as an obstacle which is untraversable --- even if the platform is actually able to pass through. Semantic understanding is important in such environments to determine which terrain is actually passable for a particular robot.

Existing approaches, which build upon deep neural models for semantic segmentation \citep{Maturana2017} or anomaly detection \citep{Wellhausen2020}, have demonstrated navigation in off-road environments; however there are recurring problems with the  collection and labelling of large amounts of relevant training data. In addition to the effort required to curate these datasets, specific class labels (tree, branch, bush, grass) do not directly correspond to the capabilities of the robotic platform.

Self-supervised systems have addressed this challenge by generating labeled datasets from past robot deployments, using classification carried out in hindsight~\citep{Wellhausen2019} or using predictions of the robot motion~\citep{Gasparino2022}. Nevertheless, these previous methods are still trained on robot-specific datasets and subsequently deployed without further adaptation. If they were to be tested in a new environment or on a new robotic platform, new data would need to be collected and the models would need to be retrained, limiting applicability. 

Achieving online self-supervision and learning are key to overcoming these aforementioned limitations, as traversability could be more easily learned in the field. 
The \gls{lagr} program~\citep{Kim2006, Hadsell2009} pioneered this direction, generating supervision during the mission and training machine learning models for traversability estimation on the fly. They showed first demonstrations of autonomous robots navigating off-road environments under this approach. 

In this work, we build upon such advances and present a system for vision-based traversability estimation that achieves online, self-supervised adaptation, by improving on various components of previous works. We name the approach \glsfirst{wvn}, as it is capable of learning which terrain is traversable by a robot after a few minutes of manual demonstrations \emph{in the wild}. The system builds upon four core ideas that we consider the key contributions of our work:
\begin{itemize}
    \item \textbf{A self-supervision system} designed for real-time operation, which concurrently generates supervision signals from vision and traversability measurements from proprioception and control performance.
    \item \textbf{A learning approach} that leverages high-dimensional, self-supervised visual features extracted using pre-trained vision transformer models, which are fed into a small neural network and efficiently trained online.
    \item \textbf{A new formulation for traversability estimation} that combines supervised learning with anomaly detection, accounting for uncertainties due to the sparse supervision. 
    \item \textbf{Closed-loop integration} with local mapping and planning methods which demonstrate that these traversability estimates are suitable for autonomous navigation tasks.
\end{itemize}
We have extensively validated our approach with ablation studies which compare against similar approaches that are trained in an offline fashion. Further, we deployed our system on real hardware, the ANYbotics ANYmal C robot, showing that it can be easily trained for navigation tasks in environments where it would not be possible to traverse using geometric mapping alone. We demonstrate fast adaptation within minutes to determine the traversable areas in a natural environment, achieving closed-loop operation in a forest with different understory foliage and terrain, and a kilometer-scale path-following task in a park where the semantic class of the path was not explicitly labeled. While our experimental results have been demonstrated on a legged robot, the principles we present are general and applicable to other ground platforms.

%% file: chapters/2_related_work.tex
\subsection{Traversability from geometry}
Classical approaches for traversability estimation analyze the geometry of the environment using 3D sensing~\citep{Moravec1985}. Recent examples from the DARPA SubT Challenge~\cite{Chung2023}, used different representations such as point clouds~\citep{Cao2022} and meshes~\citep{Hudson2022} to evaluate navigational metrics such as risk or stepping difficulty~\citep{Fan2021}.

However, a purely geometric analysis is not sufficient to completely capture traversability for a given platform. Data-driven methods can bridge this gap by learning platform-specific traversability from data or simulations. For example, \citet{Chavez-Garcia2018} learned traversability from simulations of a ground robot moving on an elevation map. \citet{Yang2021} extended this approach for legged platforms, capturing the risk of failure, energy cost and time required for navigation. Recently, \citet{Frey2022} expanded this approach to volumetric data and massive parallelization in data collection from simulation. While we recognize the contribution those previous works make in using robot-specific geometric data to determine traversability, using geometry-only does not succeed in natural environments containing natural growth such as high grass, branches or bushes. We instead focus on vision-based methods herein, as they describe semantic features that are challenging to capture otherwise.

\subsection{Traversability from semantics}
Semantic segmentation methods have been proposed to address the aforementioned challenges by assigning navigation costs to the different semantic classes. \citet{Bradley2015} presented a scene understanding system trained and evaluated using geographically diverse data. \citet{Maturana2017} demonstrated autonomous off-road navigation using semantics projected onto 3D map around a wheeled platform. \citet{Schilling2017} used semantically segmented features that were classified into fixed classes using a random forest. Recently \citet{Shaban2022} presented an approach for off-road navigation that learns a dense traversability map from sparse point-clouds. 

We note that these methods typically rely on pre-trained or fine-tuned semantic segmentation models. This requires specific class labels to be defined; and these semantic classifiers can be difficult to reuse in different environments. Nevertheless, new advances in self-supervised models, such as \dino{}~\citep{Caron2021}, are able to segment semantically meaningful classes without manual supervision. We exploit these promising tools in this work.

\subsection{Traversability from self-supervision}
Purely semantic methods are challenging to use \textit{in the wild} because (1) it is difficult to represent the relevant classes of these operating environments without labeled data, and (2) assigning a traversability cost to each class, which can also be arbitrary. \citet{Kim2006} exploited information about the areas visited by a wheeled robot to determine positive traversable samples, while a bumper was used for negative labeling. \citet{Bajracharya2009} used a similar approach combined with a height heuristic to determine long-range negative samples.

Modern methods rely on deep-learned models trained from weakly supervised data, and the supervision strongly depends on the hardware platform. \citet{Wellhausen2019} used the reprojected footholds from a legged robot to provide supervision; \citet{Zurn2021} exploited sounds produced by the platform moving on different terrain as a proxy for supervision; \citet{Gasparino2022} instead used the receding-horizon trajectory of a \gls{mpc}. Recently, TerraPN~\citep{Sathyamoorthy2022} used odometry and IMU signals as supervision and could learn a traversability model in $\SI{25}{\min}$ -- including data collection and learning. Our work is inspired by these approaches and follows similar self-supervision strategies but we aim for concurrent supervision signal generation and learning achieving orders of magnitude faster adaptation.

\subsection{Traversability from anomalies}
Anomaly detection methods are motivated by one of the key challenges of using self-supervised approaches, namely an imbalance in the number of positive and negative samples. Instead of training a discriminative model of traversability, they focus on learning generative models of the traversed terrain. This distribution can then used as a proxy to set \gls{ood} inputs as being untraversable.
\citet{Richter17} developed a visual navigation system that relied on an autoencoder to predict \gls{ood} scenes from images, switching to safer navigation behaviors when traversing novel environments.
\citet{Wellhausen2020} used multi-modal sensing from haptics, vision and depth to classify as anomalies visual elements such as flames and water reflections in navigation tasks. \citet{Ji2022} formulated a proactive anomaly detection approach that evaluated  candidate trajectories for local planning depending on their probability of failure. 
While we do not explicitly use anomalies to determine traversability, we do use it as a confidence metric to leverage the sparse supervision signals. Similar ideas have been recently presented by \citet{Seo2023} to determine traversability with point cloud data.

\subsection{Traversability from demonstrations}
\gls{irl} is a framework which can learn a reward function from demonstrations. In our context this can be interpreted as a traversability that encodes the preference to navigate certain areas. \citet{Ratliff2006} showed how IRL methods could be used to generate large-scale mission plans from aerial images. Later, \citet{Wulfmeier2017} learned cost maps to encode driving preferences using deep neural networks, which was extended by \citet{Gan2022} to guide the navigation of a legged robot in natural environments using a local reward map. While our approach uses the examples from a human operator in our self-supervised framework instead of \gls{irl}; we show that it can also be used to encode preferences depending on the demonstration and representation of the terrain.
\subsection{Adaptive traversability estimation}
A few works have demonstrated adaptive traversability estimation and navigation behavior. As part of \gls{lagr}, \citet{Kim2006} ran a clustering algorithm during deployment, and assigned positive and negative labels to the clusters through interactions to determine the traversable areas. \citet{Hadsell2009} followed a similar approach to learn different binary classifiers during exploration, using features learned from a \gls{cnn}, making it one of the first works to use neural models for this purpose. Later \citet{Lee2017} used Bayesian clustering on SLIC superpixels using color and texture features supervised by motion priors; this approach was able to propagate the traversability labels to similar segments during operation. More recently, BADGR~\cite{Kahn2021} presented an end-to-end policy designed to adapt from experiences, though it was not framed for online learning.

Our approach builds upon the \gls{lagr} ideas for self-supervision and online learning but we extend them with more powerful semantic features from a pre-trained visual transformer that allowed us to deploy our system in a variety of natural environments.

%% file: chapters/3_method.tex
\subsection{System Overview}
\begin{figure}[t]
    \centering
    \includegraphics[width=\columnwidth]{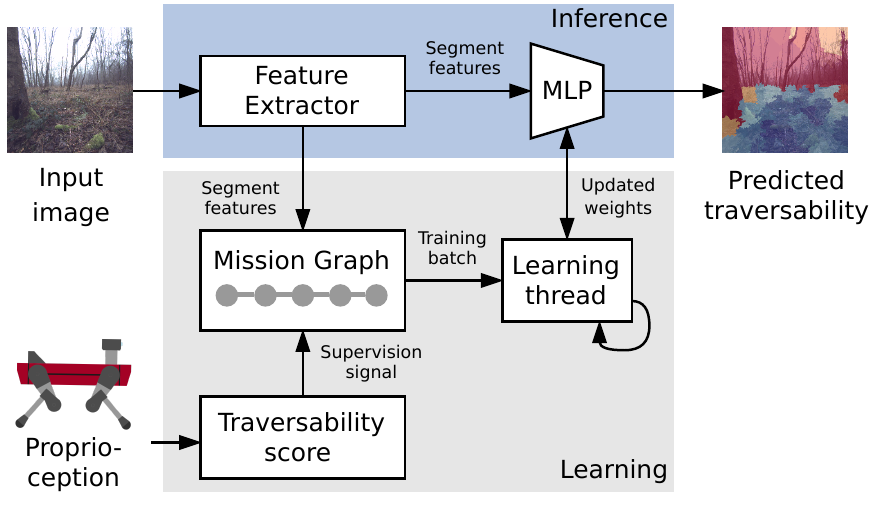}
    \caption{System overview: \gls{wvn} only requires monocular RGB images and proprioceptive data as input, which are processed to extract features and supervision signals used for online learning and inference of traversability (see \secref{sec:Method}).}
    \label{fig:block-diagram}
\end{figure}
\begin{table}
    \centering
    \ra{1.1}
    \footnotesize
    \begin{tabular}{c l}\toprule
        Symbol & Definition \\ \midrule
        $\img{}$ & RGB image with height $H$ and width $W$ \\
        $\feat{}$ & Feature map with dimensions $\dimsss{E}{H}{W}$, $E=90$ \\
        $\segmask{}$ & Weak segmentation mask with height $H$ and width $W$ \\
        $\auxsupimage{}$ &  Reprojected supervision with dimensions $\dimss{H}{W}\in [0,1]$ \\
        $\trav{}$ & Traversability score $\in [0,1]$ \\
        $\embed{n}$ & Per-segment embedding of dimension $E=90$ \\
        $\trav{n}$ & Per-segment traversability score \\
        \bottomrule
    \end{tabular}
    \caption{Main definitions used in this work}
    \label{tab:notation}
\end{table}
The objective of this work is to design a navigation system that estimates dense traversability from RGB images using a neural network model learned online, in a self-supervised manner, using labels generated by a robot interacting with its environment. We target a system which requires only a brief demonstration from a human operator for data collection and learning.

Our proposed system \gls{wvn}, builds upon different modules shown in \figref{fig:block-diagram}. \emph{Feature extraction} (\secref{subsec:feature-extraction}) and \emph{traversability score generation} (\secref{subsec:traversability-score-generation}) process the incoming sensor data to extract features and traversability scores. The \emph{mission and supervision graphs} (\secref{subsec:sup-mission-graphs}) manage the data and generate the self-supervision signals, while the \emph{learning thread} (\secref{subsec:trav-learning}) performs online traversability and anomaly learning. 
The interaction between the different modules is illustrated in \figref{fig:block-diagram}, and their implementation is covered in the following sections. The main definitions used in the rest of the paper are summarized in \tabref{tab:notation}.

\subsection{Feature Extraction}
\label{subsec:feature-extraction}
Given an RGB image $\img{}$, we first extract dense, pixel-wise visual feature maps (\emph{embeddings}) $\feat{}$. In contrast to previous works based on fine-tuned \gls{cnn}s, we rely on recent self-supervised network architectures to leverage a \gls{vit} trained using the DINO method~\citep{Caron2021} -- \dino{}. These learned representations have been demonstrated to encode meaningful semantic and instance information without requiring any labels.

\begin{figure}[t]
    \centering
    \includegraphics[width=1.0\columnwidth]{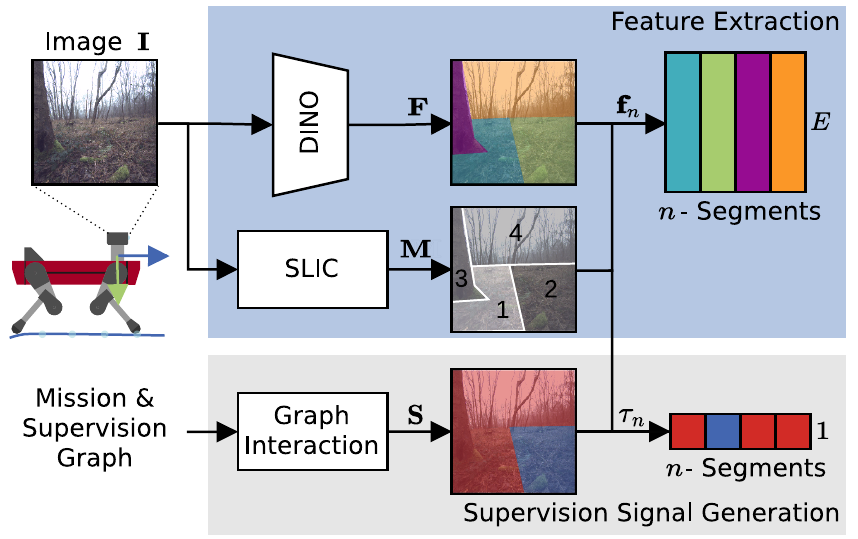}
    \caption{Feature Extraction: Our approach extracts dense $\dino{}$ features $\feat{}$ from an RGB image $\img{}$, and $n=100$ segments $\segmask{}$ using SLIC. For each of the $n$ segments we average the corresponding features to obtain per-segment embeddings $\embed{n}$. A traversability score $\trav{n}$ is computed for each segment based on the score resulting from the graph interaction process (for more detail refer to \secref{subsec:sup-mission-graphs}).}
    \label{fig:feature-extraction}
\end{figure}

Since we aim for real-time operation, the processing and storage of the full dense features on a mobile robot is prohibited by the limited compute and storage availability. Instead, we follow previous works~\cite{Lee2017} and compute a weak segmentation mask $\segmask{}$ of the input image $\img{}$ using superpixels. We use SLIC~\citep{Achanta2012} to extract 100 segments per image. We then average the feature maps segment-wise resulting in a single embedding $\embed{n}$ per segment. \figref{fig:feature-extraction} illustrates the complete feature extraction process.

\subsection{Traversability Score Generation}
\label{subsec:traversability-score-generation}
Defining which terrain is traversable or not depends on the capabilities of the specific platform. We define a continuous \emph{traversability score} $\trav{} \in [0,1]$, where $0$ is untraversable and $1$ fully traversable. Previous works have used ground reaction forces, audio supervision, or predictions from a \gls{mpc} as a proxy for this score. Instead, we adopt a simpler approach that uses the discrepancy between the robot's current linear $(x,y)$ velocity as estimated by the robot $\vel{}$, and the reference velocity command $\velcmd{}$ given by an external human operator or planning systems. The intuition is that when the robot moves on terrain that is easily traversable it should closely track the reference command. In contrast, if the robot struggles to track the reference,  the discrepancy will grow, and we can interpret it as a less traversable terrain. We define the mean squared velocity error as:
\begin{equation}
    \label{eq:traversability}
    \veldiff{} = \frac{1}{2} \left( \left( \speedcmd{x} -  \speed{x}  \right)^{2} + \left( \speedcmd{y} -  \speed{y}  \right)^{2} \right) \in \Rn{}
\end{equation}
As $\veldiff{}$ will be a noisy scalar signal, we smooth it with a 1-D Kalman Filter before passing it through a sigmoid function to obtain a valid traversability score:
\begin{equation}
    \label{eq:traversability}
    \trav{} = \sigmoid{}{ - k\left( \veldiff{} - v_{\text{thr}} \right) }
\end{equation}
with $k$ the steepness of the sigmoid, and $v_{\text{thr}}$ the midpoint of the sigmoid that assigns a traversability score of $0.5$. These values can be calibrated depending on the motion specifications of each platform and determine how the velocity error is stretched to the $[0,1]$ interval.

\begin{figure}[t]
    \centering
    \includegraphics[width=\columnwidth]{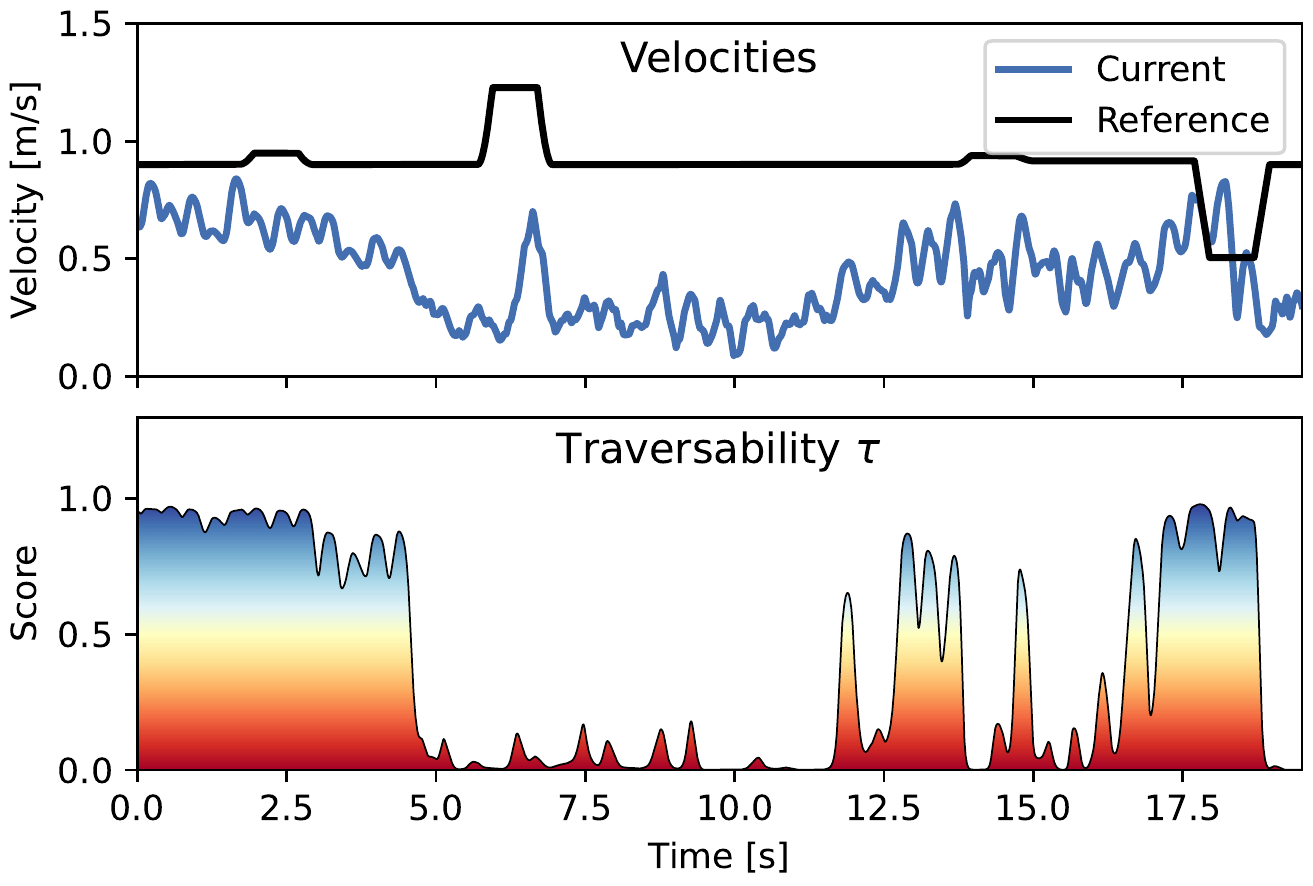}
    \caption{Traversability score generation: We compute a score based on the difference of an (external) reference velocity command and the current velocity estimated by the robot. Closely tracking the reference is interpreted as moving over traversable terrain (in blue \travsquare{}), high discrepancies indicate otherwise (red \untravsquare{}).}
    \label{fig:traversability-score}
\end{figure}

\subsection{Supervision and Mission Graphs}
\label{subsec:sup-mission-graphs}

\begin{figure*}[t]
    \centering
    \includegraphics[width=\textwidth]{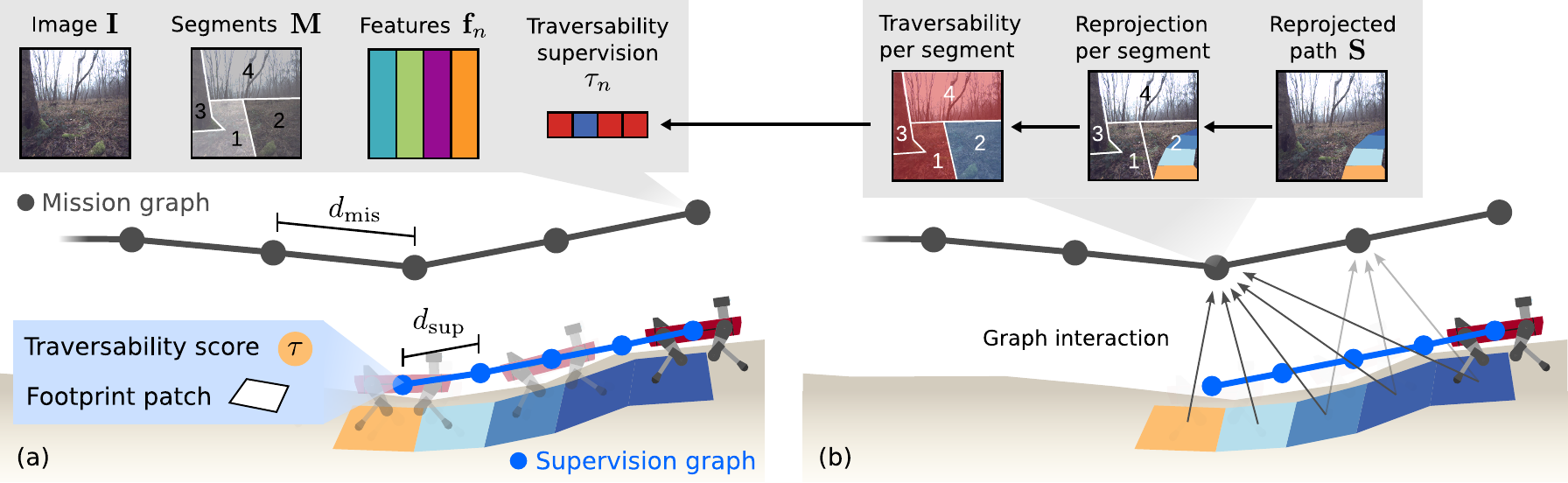}
    \caption{Supervision and mission graphs: (a) Information stored in each graph over the mission. While the Supervision Graph only stores temporary information about the robot's footprint in a sliding window, the Mission Graph saves the data required for online learning over the full mission. The color of the footprint patches indicates the generated traversability score. 
    (b) The interaction between graphs updates the traversability in the mission nodes by reprojecting the robot's footprint and traversability scores.}
    \label{fig:supervision-and-mission-graphs}
\end{figure*}

In contrast to other methods that generate the supervision signal in post-processing~\cite{Wellhausen2019, Gasparino2022, Sathyamoorthy2022}, we execute this process online by accumulating information about the recent history of operation. Our approach is inspired by graphical SLAM pipelines that often leverage both local and global graphs to integrate measurements: we maintain a \emph{Supervision Graph} to store short-horizon traversability data, and a global \emph{Mission Graph} which stores the generated training data during a mission, shown in \figref{fig:supervision-and-mission-graphs}.

\subsubsection{Supervision Graph}
The supervision graph stores within its nodes information about the current time, robot pose, and estimated traversability score (\secref{subsec:traversability-score-generation}). This graph is implemented as a ring buffer, which only keeps a fixed number of nodes $\nsupg{}$, separated from each other by a distance distance $\dsupg{}$. The product $\nsupg{}\, \dsupg{}$ determines the maximum length (in physical distance) of the supervision graph.

The stored information is a footprint track with traversability scores $\trav{}$, used to generate supervision signals in hindsight that are reprojected onto previous camera viewpoints as it is explained in the following sections.

\subsubsection{Mission Graph}
On the other hand, the mission graph stores all the information required for learning over the mission. The mission nodes are added to the graph after feature extraction if the distance w.r.t the last added node is larger than $\dmisg{}$. Each mission node contains the RGB image $\img{}$, the weak segmentation mask $\segmask{}$ and per-segment features $\embed{n}$ with their corresponding traversability supervision $\trav{n}$.

\subsubsection{Graph interaction for supervision generation}
When a new mission node is added, it triggers an interaction between the graphs to update the supervision labels $\trav{n}$ stored in each mission node (\figref{fig:supervision-and-mission-graphs}b). We reproject the footprint track and corresponding traversability scores $\trav{}$ onto all the images of the mission nodes that are within the range of the supervision graph. This ensures the information we reproject stays locally consistent in spite of potential state estimator drift.

Each mission node then has an auxiliary image with the reprojected path, $\auxsupimage{}$. We use the weak segmentation mask $\segmask{}$ to assign per-segment traversability supervision values $\trav{n}$ by averaging the score over each segment. Segments that do not overlap with the reprojected footprint track are set to zero (i.e untraversable). Then we obtain pairs of per-segment features $\embed{n}$ and traversability score $\trav{n}$ for each mission node ready for training. 

\figref{fig:supervision-and-mission-graphs} illustrates the supervision and mission graphs, and their interaction to generate supervision signals online.
%
\subsection{Traversability and Anomaly Learning}
\label{subsec:trav-learning}

We aim to train a small neural network in an online fashion that determines the feature traversability score $\trav{n}$ from a given segment feature $\embed{n}$. This allows us to predict the traversability of the scene in front of the robot from a full image by inferring only about those segments. Additionally, we explicitly model the uncertainty about the unvisited (and hence, unlabeled) areas by using anomaly detection techniques to bootstrap a confidence estimate. Our formulation also deals with non-stationary data distributions induced by continuously updating the training data and model weights. 

First, we will elaborate on how a confidence score for a segment is obtained; and then we will describe the traversability estimation task which takes as input the confidence and is jointly trained. 
\subsubsection{Confidence Estimation}
\label{subsubsec:confidence-estimation}
\begin{figure}[t]
    \centering
    \includegraphics[width=1.0\columnwidth]{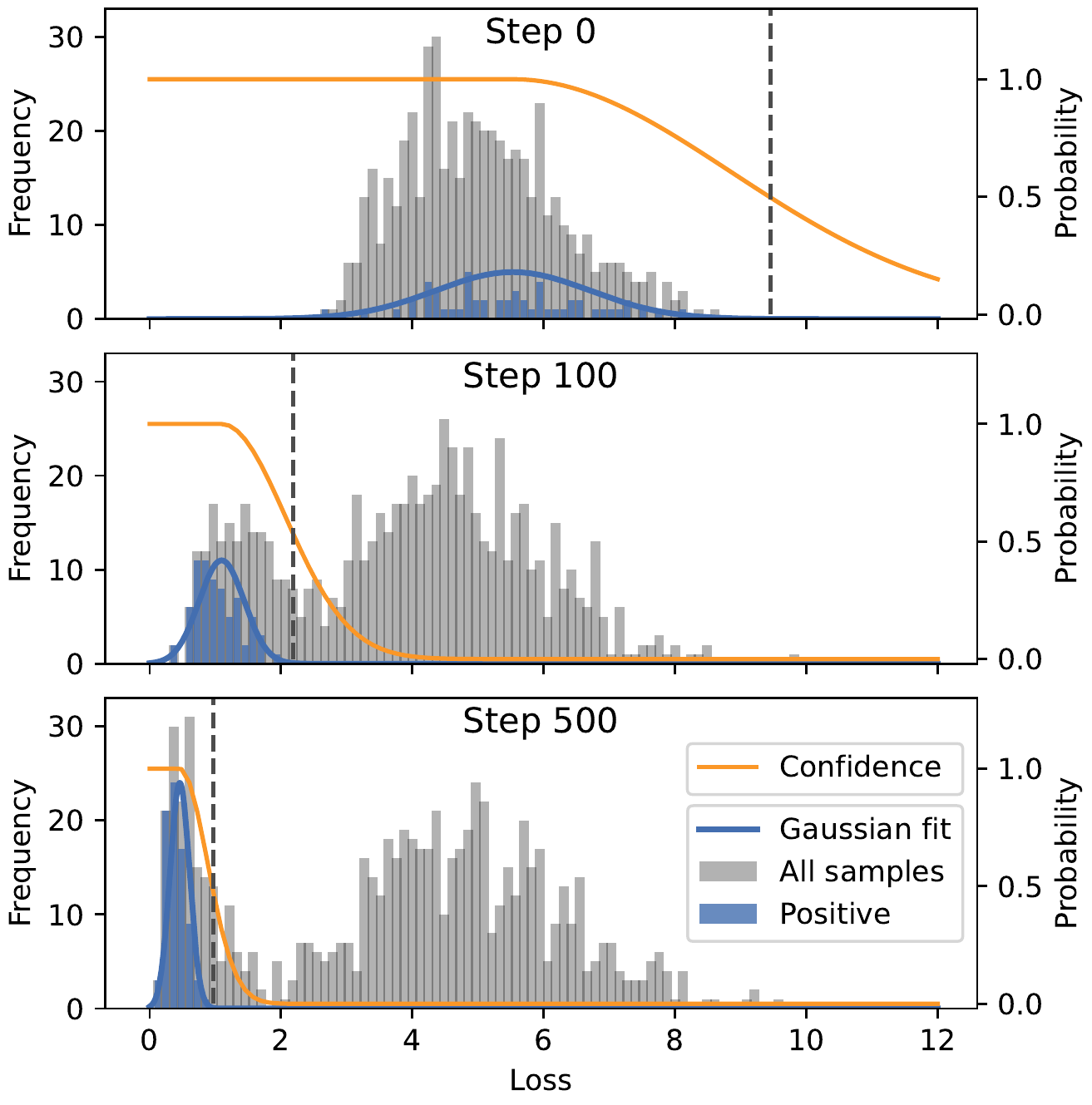}
    \caption{Histogram of the reconstruction loss $\mathcal{L}_{\text{reco}}$ distribution for all segments within a batch at optimization step 0, 100, and 500.  Traversed segments (positive samples) are shown in blue and non-traversed segments in grey. We observed that the total loss distribution becomes bi-modal as the training develops, which we use to determine a threshold to scale the confidence estimate (see \secref{subsubsec:confidence-estimation}). 
    At 500 steps all segments with a reconstruction loss $\mathcal{L}_{\text{reco}}$ over 2 are identified as fully anomalous (unconfident) and therefore induce a high loss in the traversability score when identified as traversable. The vertical grey dashed line indicates the decision boundary if a segment is detected as traversable or untraversable by the anomaly detection.}
    \label{fig:confidence-gaussian}
\end{figure}

To obtain a segment-wise confidence estimate we aim to learn the distribution over all traversed segment features $\embed{n}$. A encoder-decoder network $f_{\text{reco}}^{\theta_r}$ is trained to compress the segment feature $\embed{n}$ into a low dimensional latent space and consecutively reconstruct the original input features $\embed{n}$. 
The reconstruction loss is given by the \gls{mse} between the predicted features and the original feature compute over all channels $E$:
\begin{equation}
\mathcal{L}_{\text{reco}}(\embed{n}) = \delta_{\trav{n} \ne 0} \, \frac{1}{E} \sum_{e}{ \lVert f_{\text{reco}}^{\theta_r}(\embed{n,e}) - \embed{n,e}  \rVert^{2}},
\end{equation}
where $\delta_{\trav{} \ne 0}$ is $1$ if the segments feature traversability score $\trav{n}$ is not zero, and 0 otherwise; this ensures that the network only learns to reconstruct the embeddings that are labeled in an anomaly detection fashion. As a consequence, the trained network reconstructs known (\emph{positive}) feature embeddings, i.e. similar to the traversable segments, with small reconstruction loss; feature embeddings of unknown (\emph{anomalous}) segments the network was never tasked to reconstruct, such as trees or sky, induce a high reconstruction loss. Since the network is trained online, this results in a multi-modal reconstruction loss distribution that evolves over time, as shown in \figref{fig:confidence-gaussian}. 

The unbounded reconstruction loss $\mathcal{L}_{\text{reco}}$ for a segment is mapped to a confidence measure $c(\mathcal{L}_{\text{reco}}) \in [0,1]$ by first identifying the mode of the traversed segment losses. For this we fit a Gaussian distribution $\mathcal{N}(\muanomi{}, \sigmaanomi{})$ over the reconstruction losses per batch of the traversed segments (i.e, positive samples):
\begin{align}
n_{\text{trav}} &= \sum_{\embed{}} \delta_{\trav{n} \ne 0}, \\
\muanomi{} &= \frac{1}{n_{\text{trav}}} \sum_{\embed{}\,:\; \trav{n} \ne 0} \mathcal{L}_{\text{reco}}(\embed{n}), \\
\sigmaanomi{} &= \sqrt{ \frac{1}{n_{\text{trav}}} \sum_{\embed{}\,:\; \trav{n} \ne 0}  \left( \mathcal{L}_{\text{reco}}(\embed{n})-\muanomi{} \right)^2  }
\end{align}

We set the segment confidence to $1$ if the loss of the segment is smaller than $\muanomi{}$ and otherwise to the unnormalized Gaussian likelihood:
\begin{align}
c(\mathcal{L}_{\text{reco}}(\embed{n})) = & \exp{\left( \frac{ (\mathcal{L}_{\text{reco}}(\embed{n}) -\muanomi{})^2}{2(\sigmaanomi{}\,\sigmafactor{})^2} \right) },
\end{align}
where we introduce the tuning parameter $\sigmafactor{}$, which allows to scale the confidence.

\subsubsection{Traversability Estimation}
A small network $f_{\text{trav}}^{\theta_t}$ with a single channel output is trained to regress on the provided segment traversability score $\trav{}$. For the untraversed segments with unknown traversability score we follow a conservative approach and we assume it to be zero $\trav{}=0$, though we use the confidence score to scale their overall contribution. 
The loss for traversability estimation is computed using the confidence-weighted \gls{mse}:
\begin{equation*}
\begin{aligned}
\mathcal{L}_{\text{trav}}(\embed{}) = \; &\delta_{\trav{n}=0} \sum_{n} (1-c(\embed{n}))\, \lVert f_{\text{trav}}^{\theta_t}(\embed{n}) - 0 \rVert^{2} + \\ 
&\delta_{\trav{n} \neq 0} \sum_{n} \lVert f_{\text{trav}}^{\theta_t}(\embed{n}) - \trav{n} \rVert^{2}.
\end{aligned}
\end{equation*}
Effectively, for segments where a traversability score is available by interaction the \gls{mse} is computed. For unlabeled segments, the traversability is assumed to be zero but weighted based on the confidence score. Areas similar to the one traversed should be assigned a $c(\embed{})$ close to 1, therefore contributing insignificantly to the total loss. On the other hand anomaly areas (never traversed before, low $c(\embed{})$ score) induce a high loss if predicted with a high traversability score by $f_{\text{trav}}$. 

As we aim to provide the estimated traversability as input for a local planning system, it is desired to automatically define a threshold to determine the traversable and untraversable areas. We propose a strategy to select the traversability threshold $\travthr{}$ by measuring the current performance of the system in a self-supervised manner. We compute the \gls{roc} throughout training by classifying all segments with confidence under $0.5$ as negative and traversed segments as positive labels. Then, we decide on the traversability threshold only by setting the desired \gls{fpr}, though other metrics such as the Youden's index can also be used.

\subsubsection{Implementation details}
In our implementation, $f_{\text{reco}}^{\theta_r}$ and $f_{\text{trav}}^{\theta_t}$ are implemented by a two-layer \gls{mlp} with [256, 32] unit dense layers and ReLU non-linear activation functions.
Both networks share the weights of the hidden layers. 
$f_{\text{reco}}^{\theta_r}$ has a reconstruction head with $E$ output neurons and $f_{\text{trav}}^{\theta_t}$ a single channel traversability head followed by a sigmoid activation. 
The 32-channel hidden layer functions as the bottleneck of the encoder-decoder structure.
The total loss per segment during training is given by:
\begin{equation}
\mathcal{L}_{\text{total}}(\embed{}) = w_{\text{trav}}\mathcal{L}_{\text{trav}}(\embed{})  + w_{\text{reco}}\, \mathcal{L}_{\text{reco}}(\embed{}) .
\end{equation}
with $w_{\text{trav}}$ and $w_{\text{reco}}$ allowing to weigh the traversability and reconstruction loss respectively. We used Adam~\citep{Kingma2015} to jointly train the networks with a fixed constant learning rate of $0.001$. For a single update step, 8 valid mission nodes are randomly chosen to form a data batch; we defined a node as valid if at least a single segment of the node has non-zero traversability score. For all our experiments we set $\sigmafactor{}=2$, $w_{\text{trav}}=0.03$, $w_{\text{reco}}=0.5$ and use a maximum \gls{fpr} of $0.15$ to determine the traversability threshold. 

In \secref{subsec:ablation_study}, we present and evaluate different design choices for learning methods, features and architectures. We also evaluate the advantages of our loss formulation compared to other common approaches for traversability, such as anomaly detection (learning a distribution over positive samples) and regression on the traversability score without modeling the uncertainty.

\begin{figure}[t]
    \centering
    \includegraphics[width=1.0\columnwidth]{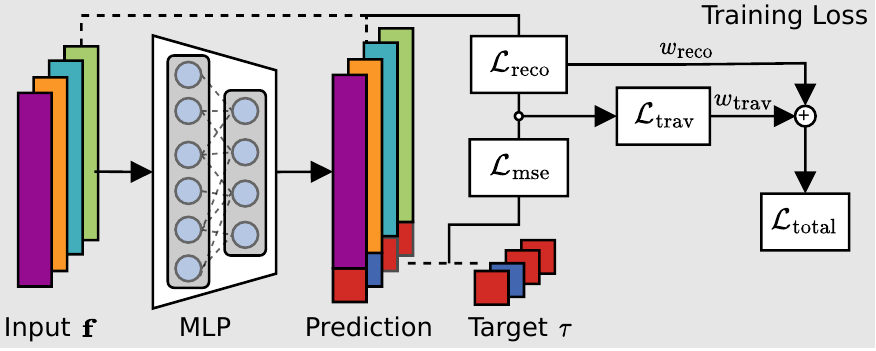}
    \caption{Network architecture and per-segment losses used for online training.}
    \label{fig:network-losses}
\end{figure}

%% file: chapters/4_closed_loop_integration.tex
We integrated the learned traversability into a standard navigation pipeline to achieve autonomous navigation with a quadrupedal platform. The details of each module of the system are explained in the following sections.

\subsection{Local terrain mapping}
\label{subsec:local-terrain-mapping}
To map the environment surrounding the robot we used an open-source terrain mapping framework~\citep{Miki2022b} to efficiently obtain a robot-centric 2.5D elevation map from the onboard depth cameras and LiDAR sensing. 
We extended this framework in~\citet{Erni2023} to fuse our predicted traversability into the local map representation. As our predicted traversability is an image, we used raycasting to take into account the occlusions with the terrain, establishing correspondences between pixel-wise traversability values in the image plane and the local map's grid cells.
This procedure allows for temporal fusion of the traversability information in the map while preserving the values of previously observed areas via exponential averaging.
Since this indirectly uses geometry, the projection method is susceptible to artifacts due to spikes in the elevation map. 

\subsection{Local planning}
\label{subsec:local-planning}
We used the projected visual traversability from the local map as a costmap for local planning. A median filter removed undesired noise and artifacts before the distance transform method ~\citep{Felzenszwalb2012} was used to obtain a \gls{sdf}, which represents the distance to the closest untraversable object. We implemented a local planner method based on \citet{Mattamala2022}, which exploits the \gls{sdf} and the local goal to generate a $\SEtwo{}$ twist command which drives the robot towards the goal while avoiding untraversable terrain. Finally, the twist command becomes the input to a robust learning-based locomotion controller based on the work by \citet{Miki2022a}, which is able to traverse rough terrain typically inaccessible to wheeled robots.

\subsection{Smart carrot for autonomous exploration}
\label{subsec:carrot}
Lastly, to generate an autonomous navigation behavior we implemented a simple exploration strategy by analyzing the robot-centric \gls{sdf} created by the local planner, by choosing a \emph{moving carrot} that drives the robot forward.

The goal pose is given by analyzing a section of the \gls{sdf} in front of the robot and selecting the position with the largest distance from all obstacles, ensuring the robot stays at the center of the traversable space. The goal is continuously updated when the \gls{sdf} is recomputed with new traversability information.
While this strategy was simple it could safely guide the robot to follow a footpath autonomously using the predicted traversability without requiring a global plan or a large-scale representation of the environment. Nevertheless, further improvement could be achieved by using a sophisticated exploration planning system.

\begin{figure}[t]
    \centering
    \includegraphics[width=1.0\columnwidth]{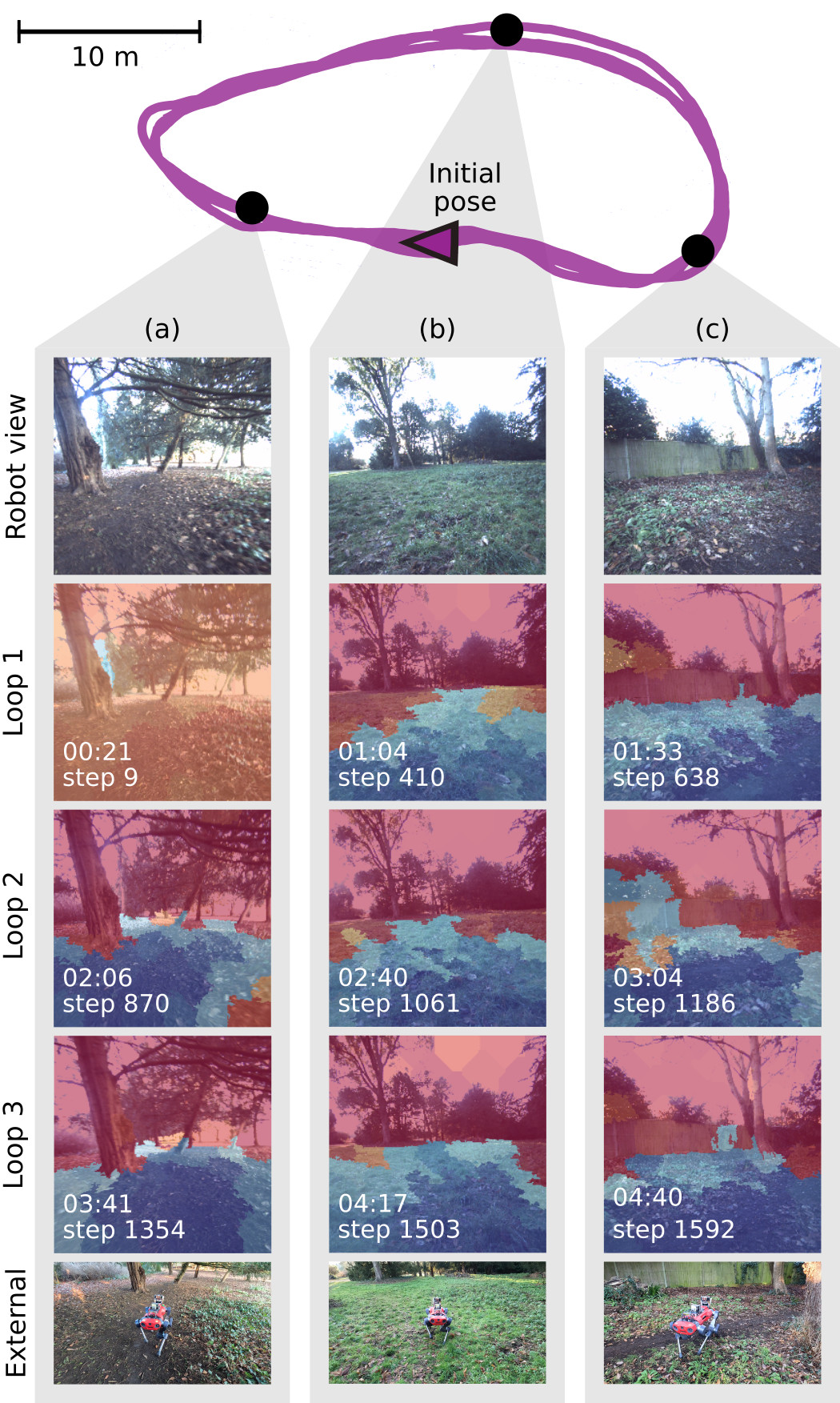}
    \caption{Adaptation on real hardware: We tested the online adaptation capabilities of our system by teleoperating the robot to complete 3 loops in a park (top, route shown in \magentasquare{}). The columns show different parts of the loop (a,b,c); each row displays the improvement of the traversability estimate over time and training steps.}
    \label{fig:loop-adaptation}
\end{figure}

%% file: chapters/5_experiments.tex
\subsection{Platform Description}
For our experiments we used an ANYbotics ANYmal~C legged robot. The robot is equipped with an additional NVidia Jetson Orin AGX to run \gls{wvn} onboard, which was implemented in pure Python code using PyTorch~\cite{Paszke2019} and ROS~1 \cite{Quigley2009}. 

The main sensing input for our system are monocular, wide \gls{fov} color images from a single global shutter Sevensense Alphasense Core camera. Additionally, the default state estimator provides $\SEthree{}$ pose and body velocity measurements. The LiDAR and depth cameras available on the robot were only used for the local terrain mapping module as described in \secref{subsec:local-terrain-mapping}.
\subsection{Real-world deployments}
We executed different deployments to validate \gls{wvn} in different environments. The first experiments highlight adaptation to new environments and the advantages of the visual traversability estimation for local planning tasks. The last two experiments demonstrate autonomous navigation among obstacles and fully autonomous large-scale path following.
%
\vspace{0.3em}
\subsubsection{Fast adaptation on hardware}
\label{subsubsec:loop-adaptation}
Our first experiment involved teleoperating the robot around 3 loops of a park environment walking on grass and dirt, on open areas and around trees. The goal was to evaluate the fast adaptation capabilities of \gls{wvn} while running on the robot.

\figref{fig:loop-adaptation} illustrates the main outcomes of the experiment, showing that the system learned to predict robot-specific traversability over the 3 loops. In particular, section \textbf{(a)} shows how the robot starts with a very poor segmentation after 9 steps of training (\SI{21}{\second}), this greatly improves after 800 steps (\SI{2}{\minute}) where it can correctly segment the dirt as traversable terrain while keeping the tree untraversable. Similar behavior occurs in section \textbf{(b)} in which the segmentation is conservative at the beginning  but it extends across the other grass patches in later iterations. Section \textbf{(c)} also illustrates some issues related to the SLIC segmentation, as some segments of the wooden wall (step 1186) are incorrectly clustered with patches of the grass, which is not observed in the other captures.

\vspace{0.3em}
\subsubsection{Benefits of visual traversability vs geometric methods}
\label{subsubsec:traversability-comparison}
\begin{figure*}[t]
    \centering
    \includegraphics[width=\textwidth]{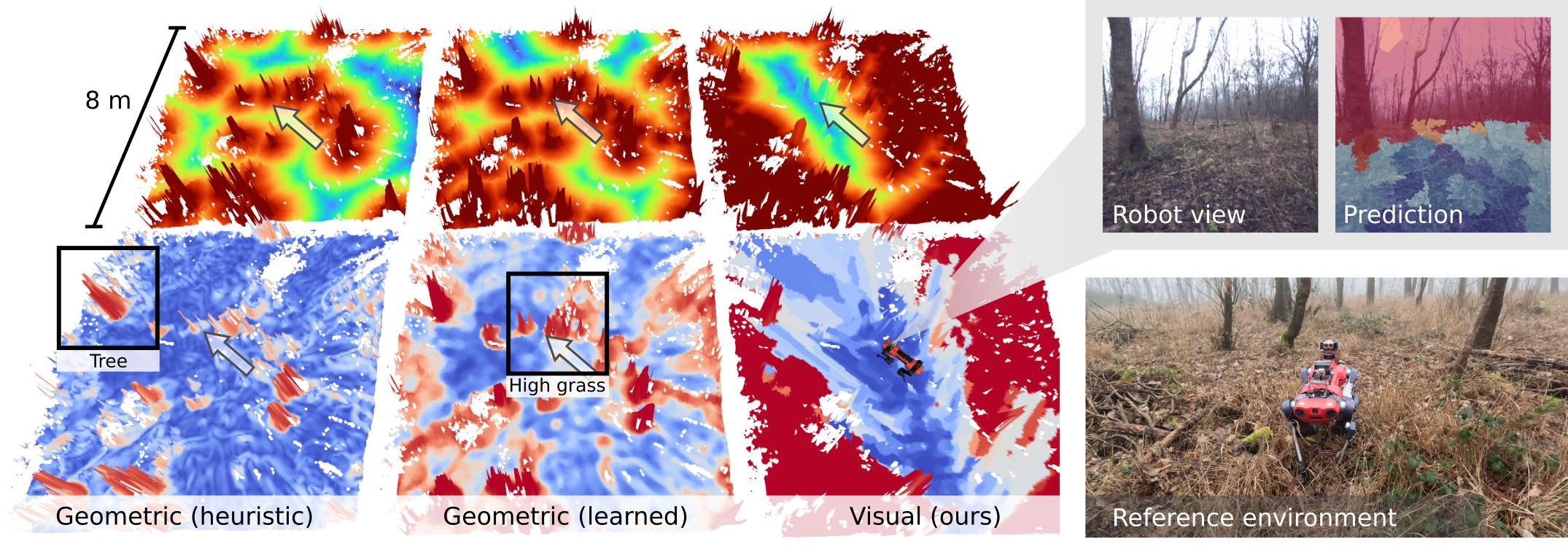}
    \caption{Visual vs geometric traversability: Illustration of traversability map (bottom row) and corresponding \gls{sdf} (top row) for three different traversability estimation methods applied to the same terrain patch. Our visual traversability estimate provides clear advantages for local planning compared to geometric methods, where the latter get heavily affected by traversable high grass or branches (bottom row). This is evident when comparing the \gls{sdf}{'s}, where geometry-based methods are more sensitive to the spikes produced by high grass areas (top row).}
    \label{fig:traversability-comparison}
\end{figure*}
Our second experiment aimed to illustrate the advantages of visual traversability estimation in challenging natural environments. Similarly to the previous experiment, we teleoperated the robot --- but in a forest with high grass, loose branches, and bushes. \figref{fig:traversability-comparison}, bottom right, shows a representative shot of the experiment, a robot's view input image and \gls{wvn}{'s} prediction, which illustrates the challenges of the environment for both vision and geometry based approaches.

To compare the different traversability methods, we used the terrain mapping module described in \secref{subsec:local-terrain-mapping}, as it allowed us to compare geometry-only and visual traversability. In particular, we compared against two geometric methods that are real-time capable and have been used in previous literature:
\begin{itemize}
\item Geometric method based on heuristics such as height and slope of the terrain~\cite{Wermelinger2016}.
\item Geometric method based on a learned model of traversability, which is part of the terrain mapping system~\cite{Miki2022b}.
\item Visual traversability provided by \gls{wvn}, raycasted onto the terrain map.
\end{itemize}
The geometric methods only require an elevation representation of the surface, and can directly determine traversability from the 2.5D geometry. For \gls{wvn} we executed a training procedure driving the robot around the environment for a few minutes first.

\figref{fig:traversability-comparison} illustrates the output \emph{traversability map} obtained by all the methods (bottom), as well as the corresponding \gls{sdf}{s} generated from them. (top) The geometric methods correctly determine the trees as untraversable areas, as they are based on the 2.5D representation. Our system is also able to successfully discriminate the trees, confirming the findings observed in \secref{subsubsec:loop-adaptation}. However, the important advantages of our method are observed in high-grass areas, which are represented as elevation spikes in the map that are classified as untraversable by the geometric approaches. \gls{wvn} correctly characterized the capabilities of the robot to successfully traverse the terrain, as demonstrated by the human operator.

When comparing the \gls{sdf}{s} such differences become more evident, as all the areas with low traversability scores become obstacles. Our system correctly determines that all the grass patches are traversable, hence the \gls{sdf}{s} displays the right classification and disregards the geometry. The only limitation of our current method is that we use just a single camera and can only predict traversability for the areas in sight. To ensure safety, unknown areas are assumed to be untraversable but future work will take advance of the robot's other cameras.
%
\vspace{0.3em}
\subsubsection{Point-to-point autonomous navigation between trees}
\begin{figure*}[h]
    \centering
    \includegraphics[width=\textwidth]{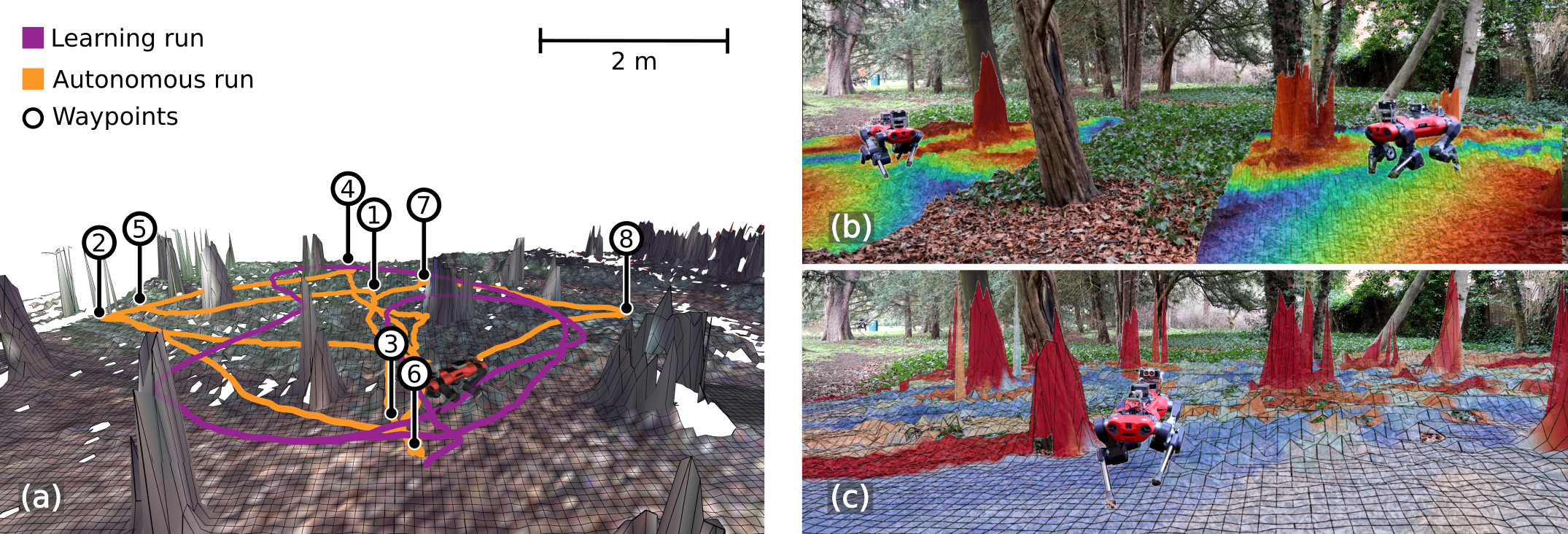}
    \caption{Point-to-point autonomous navigation: (a) After teleoperating the robot for \SI{2}{\minute} (path shown in \magentasquare{}), we successfully achieved autonomous navigation in a woodland environment (path shown in \orangesquare{}). (b) Some of the \gls{sdf}{s} generated from the predicted traversability during autonomous operation. (c) Global 2.5D reconstruction of the testing area and predicted traversability, generated in post-processing to illustrate the capabilities of our approach.}
    \label{fig:real-world-point-to-point}
\end{figure*}
After validating our approach in teleoperated settings, we executed closed-loop navigation tasks to demonstrate \gls{wvn} can easily adapt to a new environment, and the learned traversability estimate can be used to deploy the robot autonomously. 

We taught the robot to navigate in a woodland area containing dirt, high grass, and trees. A human operator drove the robot for \SI{2}{\minute} through loose dirt and grass --- an area that can be easily traversed by the legged platform. Then we commanded the local planner to execute autonomous point-to-point navigation avoiding obstacles, only using the visual traversability for closed-loop planning \secref{sec:closed-loop_integration}.

\figref{fig:real-world-point-to-point} illustrates the scene used for the experiment and the trajectories used for training and testing autonomous navigation. The robot successfully managed to reach 8 out of 8 goals, where the human operator deliberately chose targets behind trees to challenge the system. This was achieved even though neither geometry nor any additional assumptions about the environment were used during training.

We also show some examples of the \gls{sdf}{s} generated during operation used by the local planner in subfigure (b), which indicate the trees as obstacles. Lastly, in processing we fused the predicted traversability measures into a complete map in subfigure (c), which correctly aligned with the trees. However, we did observe some obstacle artifacts due to limitations of the approach, namely the use of a single camera for the predictions, the coarse segmentation from SLIC, and the raycasting process, which we further discuss in \secref{subsec:limitations}.

%
%
\vspace{0.3em}
\subsubsection{Kilometer-scale autonomous navigation in the park}
\begin{figure*}[h]
    \centering
    \includegraphics[width=1.0\textwidth]{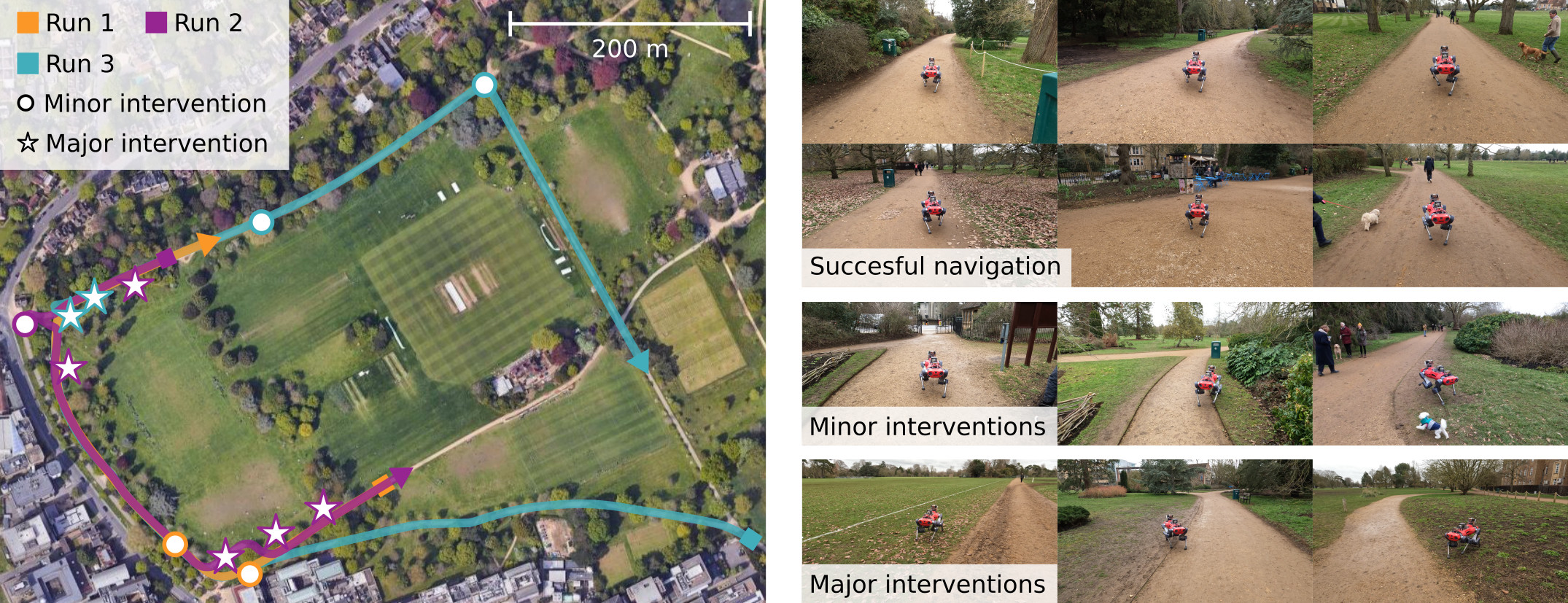}
    \caption{Kilometer-scale navigation: We deployed our system to learn to segment the footpath of a park after training for a few steps. We executed 3 runs starting from different points in the park: \orangesquare{} \emph{run 1} (\SI{0.55}{\kilo\meter}), \magentasquare{} \emph{run 2} (\SI{0.5}{\kilo\meter}), and \cyansquare{} \emph{run 3} (\SI{1.4}{\kilo\meter}). Minor interventions were applied to guide the robot in intersections; major interventions ($\mathbf{\star}$) were required for some areas when the robot miss-classified muddy patches for the path.
    }
    \label{fig:real-world-path-following}
\end{figure*}
As a last experiment, we demonstrated that \gls{wvn} can also be used to achieve preference-aware path-following behavior as a result of the human demonstrations and the online learning capabilities of the system. 

We executed 3 experiments to demonstrate this in a park. Similarly to our previous experiments, we trained the system for less than \SI{2}{\minute} along the footpath. However, we then disabled the learning thread to ensure that the predicted traversability strictly mimics the human preference during the demonstration run. The goal for the robot to follow is given by the \emph{smart carrot} module described in \secref{subsec:carrot}, which autonomously guided the robot forward along the path. 

In the 3 runs the robot was able to follow the path for hundreds of meters --- mostly staying in the center of the path, avoiding grass, bushes, benches, and pedestrians. \figref{fig:real-world-path-following} shows the trajectories followed in each run, starting from different points in the footpath. For runs $1$ and $3$ we used the same parameters, $\sigmafactor{}=2$ and \gls{fpr}$=0.15$. In run 2 we relaxed the parameters to $\sigmafactor{}=3$ and \gls{fpr}$=0.3$, producing a less conservative behavior that drove the robot to other visually similar areas in the park (very muddy grass) requiring manual intervention to correct the heading. When the robot approached an intersection we adjusted, if necessary, the heading to follow the desired footpath.

Overall, we achieved autonomous behavior that would have been difficult to achieve using only geometry, as the path boundaries were often geometrically not distinguishable. On the other hand, instead of training and using a semantic segmentation system to learn \emph{all} the possible traversable classes in the park (pavement, gravel path, roadway or grass), we showed that this short teleoperated demonstration of the gravel footpath was enough for \gls{wvn} to generate semantic cues to achieve the desired path following behavior.
%
%
%
%
\subsection{Offline validation via ablation studies}
\label{subsec:ablation_study}
To complement the results of our hardware experiments and validate our design decisions, we performed different ablation studies of the individual components of \gls{wvn}.
All the experiments were executed offline on an Nvidia RTX3080 Laptop GPU with Intel i7-11800H CPU.
%
\vspace{0.3em}
\subsubsection{Dataset overview}
\label{subsubsec:dataset_overview}
\begin{figure*}[t]
    \centering
    \includegraphics[width=1.0\textwidth]{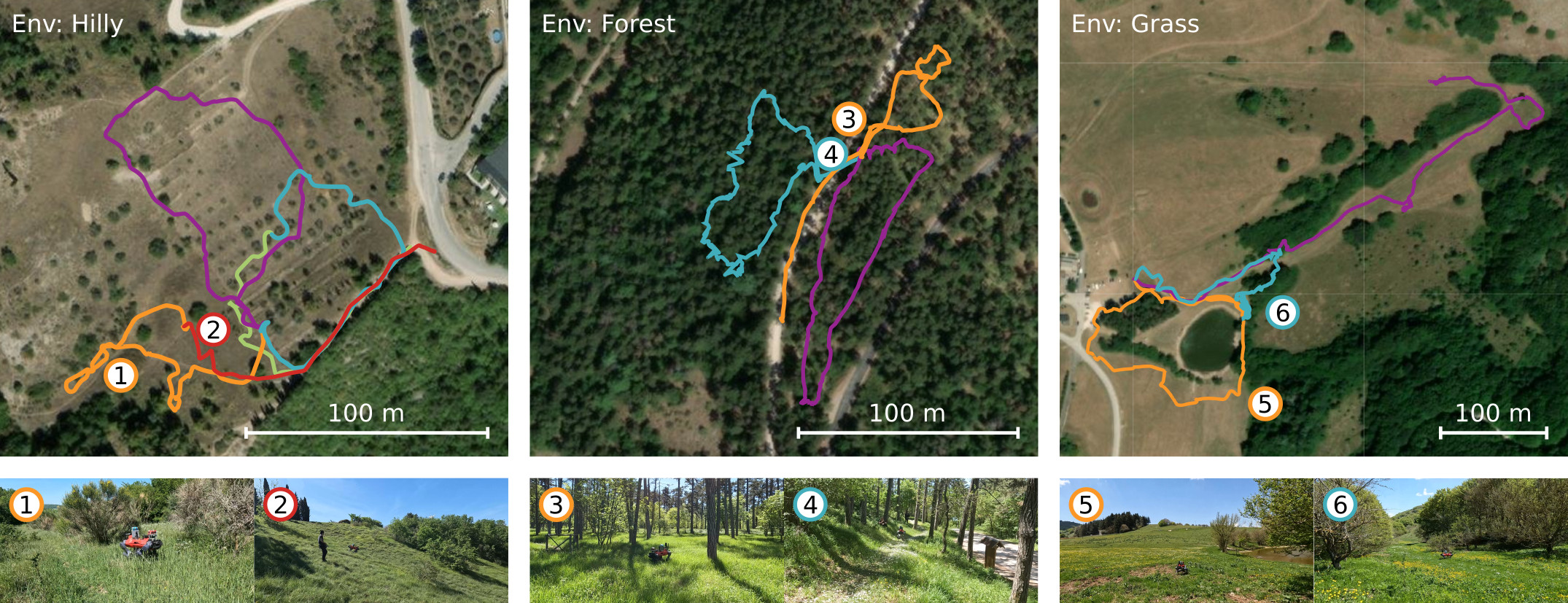}
    \caption{Aerial views of the 3 environments used for offline testing of our system, illustrating the paths used for data collection and scene examples. The purple~\magentasquare{} trajectories are used for training and the remaining for validation.}
    \label{fig:perugia-dataset}
\end{figure*}

For offline analysis we used 3 large-scale datasets of new areas not used for the real experiments:
\begin{itemize}
    \item \textit{Hilly:} a hillside with dense vegetation and fruit trees.
    \item \textit{Forest:} a fir forest with hiking paths.
    \item \textit{Grass:} a grassland area with moderate inclines and varying vegetation surrounding a small lake.
\end{itemize}
The datasets were also recorded with a teleoperated ANYmal~C platform and similar sensing setup to the previous experiments. \figref{fig:perugia-dataset} shows aerial views of the paths that were used for data collection, as well as some samples of the specific areas that were traversed during this operation.

We organized the collected data into training, validation, and testing data, which is summarized in \tabref{tab:dataset-overview}. The longest sequence recorded in each site (\figref{fig:perugia-dataset}, shown in purple~\magentasquare{}) is used for training and validation purposes. 
The first 80\% of the sequence are used to generate training data, with the remaining 20\% kept for validation. 
The remaining sequences of each scene are subsampled and exclusively used for testing. 

Regarding the labels, we manually segmented images from the test split into traversable ($1$) and untraversable ($0$) classes, which we named \emph{ground truth labels} (\emph{GT}). These binary labels reflect the intuition of an expert robot operator on which places are safely accessible for the robot, and were used for quantitative assessment of the design decisions. While simulation-based evaluations could provide a precise traversability score, we disregarded it because of the limitation of sim-to-real transfer of vision-based methods. Human supervision was straightforward and could incorporate more subtle risks and preferences.

On the other hand, we also used the labels generated by executing \gls{wvn} over the sequences using the self-supervision approach, which we name \emph{SELF}. Since the \emph{GT} labels were binary due to intrinsic challenges of producing ground truth continuous traversability signals, we also binarized the \emph{SELF} labels for a fair comparison. 

Lastly, even though we only had access to binary ground truth labels we did not change the regression formulation presented in \secref{subsec:trav-learning}, as we could also obtain a binary output by thresholding the continuous signal proposed by our system. This resulted in a classification task that could be evaluated using metrics such as \gls{accuracy}.

\begin{table}[h]
\centering
\ra{1.0}
\footnotesize
\setlength{\tabcolsep}{3pt}
\caption{Dataset Overview}
\begin{threeparttable}
\begin{tabular}{llcrccc}
\toprule
\textbf{Env}    & Split & Duration  & Distance & \# Traj & \# Image & Label  \\ \midrule 
\textbf{Hilly}   & Train & \SI{512.4}{s}    & \SI{262}{m} & 1 & 920 & \emph{SELF} \\
                & Val   & \SI{121.1}{s}    & \SI{66}{m} & 1 & 230 & \emph{SELF} \\
                & Test  & \SI{1202.2}{s}   & \SI{840}{m} & 4 & 55 & \emph{GT} \\ \midrule
\textbf{Forest\tnote{$\star$}} & Train & \SI{402.1}{s}    & \SI{606}{m} & 1 & 991 & \emph{SELF} \\
                & Val   & \SI{134.0}{s}    & \SI{151}{m} & 1 & 247 & \emph{SELF} \\
                & Test  & \SI{970.5}{s}    & \SI{896}{m} & 2 & 41 & \emph{GT} \\ \midrule
\textbf{Grass}  & Train & \SI{860.3}{s}    & \SI{857}{m} & 1 & 2050 & \emph{SELF} \\
                & Val   & \SI{242.5}{s}    & \SI{214}{m} & 1 & 512 & \emph{SELF} \\
                & Test  & \SI{2196.2}{s}   & \SI{1224}{m} & 3 & 113 & \emph{GT} \\ \bottomrule
\end{tabular}
  \begin{tablenotes}\footnotesize
    \item[$\star$] Length measured using RTK-GPS and may not reflect the real-distance traversed within the forest.
  \end{tablenotes}
\end{threeparttable}
\label{tab:dataset-overview}
\end{table}

\vspace{0.3em}
\subsubsection{Evaluation method}
For all of our experiments we trained and tested in the same environment using the splits previously presented, unless stated otherwise.
Regarding the metrics, we evaluated our system in a binary classification setting due to the limitations of the \emph{GT} labels previously discussed. Hence, we reported the \gls{accuracy} for all of our experiments. The accuracy is computed in image space (i.e. pixel-wise) as opposed to segment-wise, which accounts for misclassifications induced by image segments containing both traversable and untraversable terrain.

Lastly, all compared neural network models are trained for 1000 steps with 5 different random seeds, and we report confidence intervals for all the metrics. 
%
\vspace{0.3em}
\subsubsection{Study 1: Learning Methods}
Our first study compared our approach to \emph{classical} machine learning methods previously used for traversability estimation, namely \gls{rf}~\cite{Schilling2017} and \gls{svc}~\cite{Bajracharya2009}. For \gls{svc} we followed their work and use a polynomial kernel of degree 2 and \gls{rbf} kernel. All methods are trained using the \dino{} features. 

\tabref{tab:learning-model} summarizes the main results of this study. We observed that \emph{our} method based on an \gls{mlp} performs consistently good across all three environments with respect to the \emph{GT} labels (see \tabref{tab:learning-model}). Interestingly, the \gls{rf} is competitive and still performs strongly across scenes, specially well in the \textit{Forest} scene.
The results confirm that our chosen model overall outperforms machine learning methods previously used in the literature, while allowing us to continuously adapt the model using gradient descent during the mission. 
\begin{table}
  \centering
  \ra{1.1}
  \footnotesize
  \setlength{\tabcolsep}{2pt}
  \begin{tabular}{llcccc}\toprule
\textbf{Env}    & Metric         & RF              & SVC-RBF        & SVC-Poly               & WVN       \\ \midrule
  \textbf{Hilly} & \emph{GT}    & 71.36 $\pm$ 0.0   & 65.46 $\pm$ 0.53 & 73.96 $\pm$ 1.61 &  \textbf{81.05} $\pm$ 1.11  \\
   & \emph{SELF}                & 88.32 $\pm$ 0.0   & 87.51 $\pm$ 0.25 & 84.81 $\pm$ 0.31 & 78.58 $\pm$ 0.82  \\ \midrule
  \textbf{Forest} & \emph{GT}     & \textbf{83.07} $\pm$ 0.44 & 74.66 $\pm$ 0.55 & 76.28 $\pm$ 1.34 & \textbf{82.45} $\pm$1.10 \\
   & \emph{SELF}                & 81.95 $\pm$ 0.44 & 90.19 $\pm$ 0.20 & 88.09 $\pm$ 0.53  & 85.26 $\pm$ 1.24  \\ \midrule
  \textbf{Grass} & \emph{GT}    & 59.16 $\pm$ 0.0   & 63.64 $\pm$ 0.33 & 69.02 $\pm$ 2.00 & \textbf{78.21} $\pm$ 2.39 \\
   & \emph{SELF}                & 88.14 $\pm$ 0.0   & 87.79 $\pm$ 0.14 & 86.74 $\pm$ 0.39 & 82.60 $\pm$ 0.29\\
  \bottomrule
  \end{tabular}
  \caption{Learning Method: Traversability Accuracy of Random Forest (RF), Support Vector Classifier (SVC), and WVN with respect to the binary ground truth labels \emph{GT} and self-supervised labels \emph{SELF} on the ablation environments.}
  \label{tab:learning-model}
\end{table}
%
\subsubsection{Study 2: Training Objective}
We then studied the impact of various training objectives on the model performance. In particular, we compared four approaches:
\begin{itemize}
    \item \emph{Trav}: We trained our network to directly regress on the traversability by assuming all untraversed segments are untraversable: no reconstruction loss $w_{\text{reco}}=0$, with full confidence $c(\embed{})=1$, and without self-supervised $\travthr{}$ thresholding.
    \item \emph{Fixed threshold}: We fixed the traversability threshold $\travthr{}$ to a value of $0.5$.
    \item \emph{Anomaly detection}: We used the confidence score to classify features with a confidence over $0.5$ as traversable.
    \item \gls{wvn}: Our full method.
\end{itemize}
Our proposed method consistently outperforms the other settings (\tabref{tab:learning-objective}). A significant performance increase (+5.17\%) was achieved over the \textit{Trav} simplest traversability setting by adding the confidence-weighted traversability loss formulation (\textit{Fixed-threshold}).
Generally, in the \textit{Trav}-setting more regions are classified as untraversable leading to an over-conservative traversability estimation, which performed well on the \emph{SELF}-labels but does not reflect the \emph{GT} traversability.
Further improvement of 7.33\% can be achieved by adding the online traversability threshold scaling. Here it is important to mention that a fixed threshold is insufficient during deployment given the online adaptation of the network. Lastly, the experiment using only \textit{Anom} detection performs reasonably well, suggesting that a meaningful confidence score was learned across various environments. This indicates that the learned anomaly detection is useful for guiding the traversability learning of our method. 
\begin{table}
  \centering
  \ra{1.1}
  \footnotesize
  \setlength{\tabcolsep}{2pt}
  \begin{tabular}{llcccc}\toprule
\textbf{Env}    & Metric         &Trav           & Fixed-Threshold       & Anom            & WVN       \\ \midrule
  \textbf{Hilly} & \emph{GT} &   65.46 $\pm$ 0.53 & 73.96 $\pm$ 1.61 & 72.92 $\pm$ 0.82 & \textbf{81.05} $\pm$ 1.11  \\
   & \emph{SELF} & 87.51 $\pm$ 0.25 & 84.81 $\pm$ 0.31 & 69.21 $\pm$ 0.85 & 78.58 $\pm$ 0.82  \\ \midrule
  \textbf{Forest} & \emph{GT}  & 74.66 $\pm$ 0.55 & 76.28 $\pm$ 1.34 & 70.31 $\pm$ 1.12 & \textbf{82.45} $\pm$1.10 \\
   & \emph{SELF}& 90.19 $\pm$ 0.20 & 88.09 $\pm$ 0.53 & 68.37 $\pm$ 1.44 & 85.26 $\pm$ 1.24  \\ \midrule
  \textbf{Grass} & \emph{GT}  &  63.64 $\pm$ 0.33 & 69.02 $\pm$ 2.00 & 77.64 $\pm$ 0.44 & \textbf{78.21} $\pm$ 2.39 \\
   & \emph{SELF}&  87.79 $\pm$ 0.14 & 86.74 $\pm$ 0.39 & 75.29 $\pm$ 0.21 & 82.60 $\pm$ 0.29\\
  \bottomrule
  \end{tabular}
  \caption{Training Objective: Traversability Accuracy for different learning objectives with respect to the binary ground truth labels \emph{GT} and self-supervised labels \emph{SELF} on the ablation environments. Refer to \textit{Study 2: Training Objective} for further details.} 
  \label{tab:learning-objective}
\end{table}
%
%
%
\vspace{0.3em}
\subsubsection{Study 3: Features}
The quality of extracted segment features can have a significant impact on the performance of traversability prediction, given that we only consider a fixed feature extraction backbone. Hence, we evaluated the performance of the selected self-supervised pre-trained \dino{} backbone against popular residual network architectures pre-trained on ImageNet (ResNet-50, EfficientNet-B4), and also trained using self-supervised learning, such as ResNet-50 trained with DINO (DINO-ReN). We also compared against other classical features, such as dense SIFT~\citep{Lowe2004} features over RGB channels. 
We fed the features to the \textit{trav}-setting model introduced in the previous study, to isolate the impact of the feature extractor on the traversability estimation performance from other procedures. We present the model parameter count for each feature extractor and inference time of all network architectures on a Jetson Orin in \tabref{tab:feature_ablation}.
The features generated by methods using self-supervised pre-training clearly outperformed pre-trained models on ImageNet, even when using the same architecture, aligning with the findings of \citet{Caron2021}. The short inference time measured on the Orin board also validates the choice of \dino{} as the backbone suitable for \gls{wvn}.
\begin{table}
\centering
\ra{1.1}
\begin{tabular}{clcccc}\toprule
 & Architecture & Param  & Time & \emph{GT} & \emph{SELF}\\\midrule
\parbox[t]{1mm}{\multirow{5}{*}{\rotatebox[origin=c]{90}{\textbf{Hilly}}}} & \dino{} & 21M   & \SI{17.0}{\milli\second} & \textbf{65.46} $\pm$ 0.53 & 	87.51 $\pm$ 0.25 \\ 
 & DINO-ReN                                                               & 23.5M & \SI{15.9}{\milli\second} & 64.05 $\pm$ 0.08 & 	86.71 $\pm$ 0.03 \\ 
 & EffNet-B4                                                              & 17.5M & \SI{26.1}{\milli\second} & 62.55 $\pm$ 0.07 & 	86.09 $\pm$ 0.02 \\ 
 & ResNet-50                                                              & 23.5M & \SI{15.9}{\milli\second} & 61.87 $\pm$ 0.08 & 	85.37 $\pm$ 0.02 \\ 
 & SIFT                                                                   & 0     & - & 58.78 $\pm$ 0.73 & 	82.80 $\pm$ 0.03 \\ \midrule
\parbox[t]{1mm}{\multirow{5}{*}{\rotatebox[origin=c]{90}{\textbf{Forest}}}} & \dino{} & 21M   & \SI{17.0}{\milli\second}   & \textbf{74.66} $\pm$ 0.55 & 90.19 $\pm$ 0.20 \\ 
 & DINO-ReN                                                              & 23.5M & \SI{15.9}{\milli\second}  & 74.14 $\pm$  0.24 & 89.12 $\pm$ 0.21 \\ 
 & EffNet-B4                                                                & 17.5M & \SI{26.1}{\milli\second}  & 73.00 $\pm$  0.76 & 89.63 $\pm$ 0.29 \\ 
 & ResNet-50                                                                & 23.5M & \SI{15.9}{\milli\second}  & 72.71 $\pm$  0.22 & 88.49 $\pm$ 0.10 \\
 & SIFT                                                                  & 0     & -  & 60.94 $\pm$  0.00 & 83.27 $\pm$ 0.00 \\ \midrule
\parbox[t]{1mm}{\multirow{5}{*}{\rotatebox[origin=c]{90}{\textbf{Grass}}}} & \dino{} & 21M   & \SI{17.0}{\milli\second}   & 63.64 $\pm$ 0.33  & 87.79 $\pm$ 0.14 \\
 & DINO-ReN                                                              & 23.5M & \SI{15.9}{\milli\second}  & \textbf{67.08} $\pm$  0.36  &	88.15 $\pm$ 0.13 \\
 & EffNet-B4                                                               & 17.5M & \SI{26.1}{\milli\second}  & 61.01 $\pm$  0.78  &	85.91 $\pm$ 0.00 \\
 & ResNet-50                                                                & 23.5M & \SI{15.9}{\milli\second}  & 62.99 $\pm$  0.03  &	85.23 $\pm$ 0.12 \\
 & SIFT                                                                  & 0     & -  & 57.61 $\pm$  0.91  &	83.79 $\pm$ 0.02 \\
\bottomrule
\end{tabular}
\caption{Comparison of feature extraction backbones.}
\label{tab:feature_ablation}
\end{table}
%
\vspace{0.3em}
\subsubsection{Study 4: Scene Adaptation}
\begin{figure*}[t]
    \centering
    \includegraphics[width=1.0\textwidth]{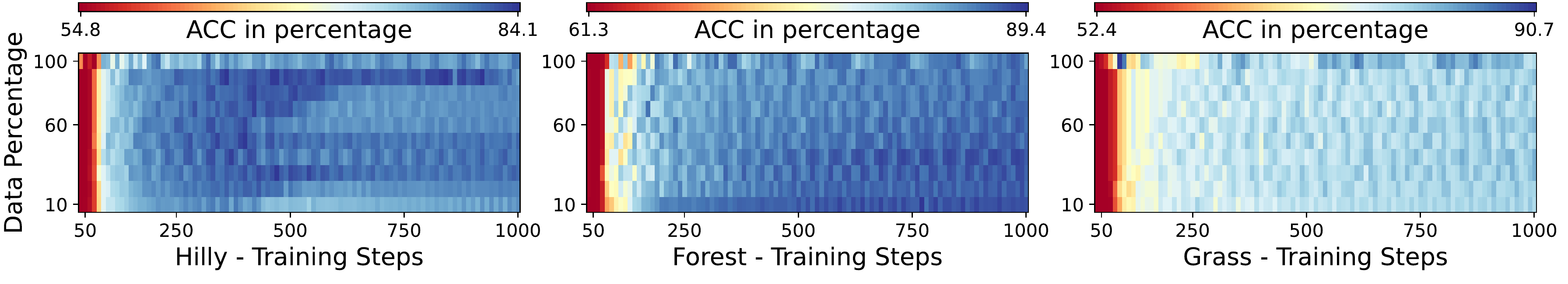}
    \caption{Adaptation Speed vs Dataset Size: The performance measured by the accuracy increases over training as expected. In the limit case of training for a few steps ($<50$), the performance is equally degraded --- independent of the dataset size. A small dataset size is sufficient for good performance. Please observe the different color scales for each environment. In the \textit{Grass} environment the color scale is distorted by an outlier when using $100$\% of the data after 70 steps.}
    \label{fig:time-data-ablation}
\end{figure*}

We evaluated the performance of \gls{wvn} when trained on one environment and tested on all the others, to test the necessity for online adaptation.
\tabref{tab:scene_ablation} shows the resulting accuracy for each scene combination. We observed that in general the best performance is achieved when testing on the same training environment as expected, dropping otherwise.
The model trained on \textit{Hilly} performs overall the best and showed specially good performance on \textit{Grass}, which we suspect is due to the visual similarity of the scenes (see \figref{fig:perugia-dataset}).

In general, we remark that even though the robot was deployed within scenes featuring similar semantic classes (e.g. trees or high grass), on the same day and within a few kilometers radius, the performance still degraded. This suggests even worse performance drops for changing seasons or urban to natural environment scene changes.
We argue that even though this can be hypothetically mitigated by increasing the amount of training data, this is costly, and online adaptation provides a practical solution to enable the deployment of robots in new or changing environments.
\begin{table}
\centering
\ra{1.1}
\begin{tabular}{cccc}\toprule
Training & \textbf{Hilly} & \textbf{Forest} & \textbf{Grass} \\\midrule
\textbf{Hilly} &   \textbf{81.05} $\pm$1.11 & 82.14 $\pm$ 1.78 & \textbf{82.14} $\pm$ 0.63 \\ 
\textbf{Forest} &  75.86 $\pm$ 2.18 & \textbf{82.45} $\pm$1.10 & 75.80 $\pm$ 2.80 \\ 
\textbf{Grass} &   77.49 $\pm$ 4.36 & 73.22          $\pm$ 6.38 & 78.21 $\pm$2.39 \\
\bottomrule
\end{tabular}
\caption{Scene Adaptation: Traversability Accuracy with respect to the \emph{GT} labels. Each row corresponds to a training run on the specific environment and testing on all environments. }
\label{tab:scene_ablation}
\end{table}
%
%
%
\vspace{0.3em}
\subsubsection{Study 5: Adaptation Speed \& Dataset Size}
\begin{figure*}[t]
    \centering
    \includegraphics[width=1.0\textwidth]{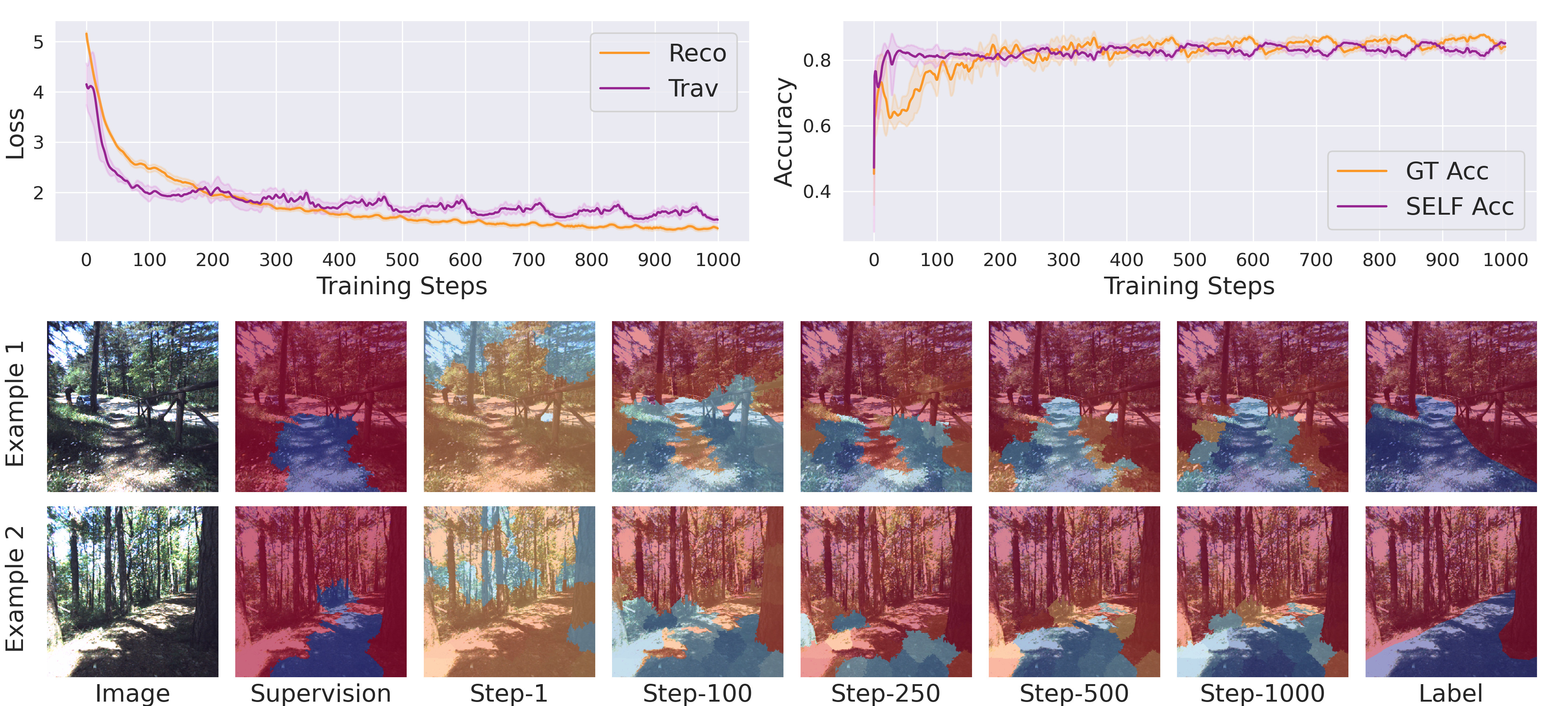}
    \caption{Training Process: We detail the incremental training process executed by \gls{wvn} in terms of the loss (top left), accuracy (top right), and visual examples (bottom).}
    \label{fig:learning_speed}
\end{figure*}
For our final study we investigated how fast can \gls{wvn} adapt to the new environments and how many data samples are needed. 
To examine this we designed an experiment in which we trained the network for 1000 steps for different training dataset sizes, ranging from $10$\% to $100$\% of the original size. We measured the accuracy each $2$nd step and every $10$\% increment of the dataset size.

As a result, we obtained heatmaps displaying the performance evolution across these 2 variables, shown in \figref{fig:time-data-ablation}. For all environments starting from a randomly initialized network, we observed that good performance can be achieved within 200 steps. We argue that this is due to use of segments: adding a single image provides 100 new training samples for the network.
During continuous training, we also observed some slight fluctuation with respect to the test accuracy. \figref{fig:learning_speed} shows the training loss accuracy over time, illustrating some of the fluctuating behavior, as well as example images of the output segmentation over training.
%
\subsection{Limitations}
\label{subsec:limitations}
We have demonstrated that \gls{wvn} can learned a traversability estimate online, allowing immediate deployment of robots in new environments. However, we observed some limitations during our ablation studies and field deployment.
\begin{itemize}
\item \gls{wvn} runs at \SI{2.5}{\hertz} on the NVidia Orin board with unoptimized code. Improving the efficiency of the pipeline would allow for faster training and better autonomous navigation performance.
\item The use of superpixels to reduce the computational complexity limits the accuracy of the output segmentation, as the SLIC segments are only similar from a color-space sense, consequently affecting the computation of features per segment by averaging semantically different features.
\item While the use of velocity tracking error was a simple proxy for the traversability score, it does not fully characterize the different interactions the robot can have with the environment. Further investigation is required to determine alternative metrics that can be more suitable for specific platforms.
\item For closed-loop integration we projected the traversability prediction onto a local terrain map, which allowed for a straightforward integration with the local planner. However, this presented important drawbacks due to perspective projection, limited \gls{fov} due to single camera usage, and raycasting on an inaccurate 2.5D map that required additional filtering stages.
\item Lastly, our system could be also framed within the continual learning paradigm. While we do not address it explicitly, we believe that \gls{wvn} could greatly benefit from the advances of continual learning --- not only to adapt within a single mission --- but between different test environments.
\end{itemize}

%% file: chapters/6_conclusion.tex
We presented \glsfirst{wvn}, a system that leverages the latest advances in pre-trained self-supervised networks with a scheme to generate supervision signals while a robot operates, to achieve online, onboard visual traversability estimation. The fast adaptation capabilities of our system allowed us to easily deploy robots for navigation tasks in new environments after just a few minutes of learning from human demonstrations. 
We validated \gls{wvn} through different ablation studies and real-world experiments, illustrating its fast adaptation capabilities, the consistency of its traversability prediction for local planning, and \SI{1.4}{\kilo\meter} closed-loop navigation experiments in natural scenes. Our experiments show that \gls{wvn} can enable autonomous robot navigation by learning from small data \emph{in the wild}. 
We aim to tackle the current limitations of our system by exploring data-driven self-supervised methods for segment extraction, possibly mitigating artifacts induced by segments containing traversable and untraversable terrain. We also plan to extend the current implementation to multiple cameras --- allowing the system to learn from different inputs and estimate traversability in different directions for more complex local planning. For future work specific to legged systems capable to negotiate challenging terrain, we aim to further close the loop between \gls{wvn}{'s} traversability prediction and feedback provided directly by the locomotion policy about the traversability of the terrain.

%% file: chapters/acknowledgments.tex
This work was supported by the Swiss National Science Foundation (SNSF) through project 188596, the National Centre of Competence in Research Robotics (NCCR Robotics), the European Union's Horizon 2020 research and innovation program under grant agreement No 101016970, No 101070405, and No 852044, and an ETH Zurich Research Grant. Jonas Frey is supported by the Max Planck ETH Center for Learning Systems. Matias Mattamala is supported by the National Agency for Research and Development (ANID) / DOCTORADO BECAS CHILE/2019 - 72200291 and NCCR Robotics. Maurice Fallon is supported by a Royal Society University Research Fellowship.